%% file: main.tex
\RequirePackage{fix-cm}
\documentclass[twocolumn]{svjour3}          
\smartqed  
\usepackage{graphicx}
\usepackage{natbib}
\usepackage{multirow}
\usepackage{amsmath}
\usepackage{amssymb}
\usepackage{bbm}
\usepackage{bm}
\usepackage{booktabs}
\usepackage{array} 
\usepackage{blindtext}
\usepackage{comment}
\usepackage{url}

\usepackage{graphicx}%
\usepackage{multirow}%
\usepackage{amsmath,amssymb,amsfonts}%

\usepackage{amsthm}%
\usepackage{mathrsfs}%
\usepackage[title]{appendix}%
\usepackage{xcolor}%
\usepackage{textcomp}%
\usepackage{manyfoot}%
\usepackage{booktabs}%
\usepackage{algorithm}%
\usepackage{algorithmicx}%
\usepackage{algpseudocode}%
\usepackage{listings}%
\usepackage{color}
\usepackage[dvipsnames, table]{xcolor}
\usepackage{pifont}
\usepackage{arydshln}
\usepackage{soul}
\usepackage{makecell}
\usepackage{wrapfig}

\newcommand{\benchname}{BuddyVQA} 
\newcommand{\modelname}{MyBuddy}

\newcommand{\cmark}{\color{green}\ding{51}}%
\newcommand{\xmark}{\color{red}\ding{55}}%

\definecolor{lightblue}{RGB}{173, 216, 230}
\definecolor{lightgreen}{RGB}{144, 238, 144}
\definecolor{lightyellow}{RGB}{255, 255, 224}
\definecolor{lightred}{RGB}{255, 182, 193}
\definecolor{lightpurple}{RGB}{216, 191, 216}
\definecolor{lightorange}{RGB}{255, 219, 187}
\definecolor{darkblue}{RGB}{123, 166, 180}   
\definecolor{darkgreen}{RGB}{94, 188, 94}    
\definecolor{darkyellow}{RGB}{245, 245, 184} 
\definecolor{darkred}{RGB}{205, 132, 143}    
\definecolor{darkpurple}{RGB}{166, 141, 166}

\newcommand{\eg}{\textit{e.g.}}
\newcommand{\ie}{\textit{i.e.}}
\newcommand{\vs}{\textit{vs. }}
\newcommand{\wrt}{\textit{w.r.t. }}

\makeatletter
\renewcommand\paragraph{
  \@startsection{paragraph} 
  {4} 
  {\z@} 
  {.5em \@plus1ex \@minus.2ex} 
  {-.5em} 
  {\normalfont\normalsize\bfseries} 
}
\makeatother
\begin{document}
\sloppy

\title{Companion-style QA Assistance in Ego-Vision}

\author{Hangyu~Qin \and
        Junbin~Xiao \and
        Shenglang~Zhang \and
        Angela~Yao
}

\institute{
Corresponding to: junbinxiao@ustc.edu.cn. \\
Hangyu Qin and Angela Yao are with National University of Singapore. 
\\
\indent Junbin Xiao and Shenglang Zhang are with University of Science and Technology of China.
}
\date{Received: date / Accepted: date}

\maketitle

\input{abstract}
\input{intro}
\input{related}
\input{dataset}

\input{method}
\input{experiment}
\input{conclusion}
\input{limitation}
\paragraph{Competing Interests}
All authors certify that they have no affiliations with or involvement in any organization or entity with any financial interest or non-financial interest in the subject matter or materials discussed in this manuscript.

\paragraph{Data Availability Statement}
The authors confirm that this manuscript has associated data in a data repository. The data supporting the findings of this work will be made publicly available. Specifically, this includes the \benchname\ dataset and the \modelname\ method. Instructions for access will be provided to ensure reproducibility.

\bibliographystyle{spbasic}      
\bibliography{main}   

\clearpage
\appendix
\input{appendix}
\end{document}

%% file: abstract.tex
\begin{abstract}
AI companions are envisioned as always-on assistants that support users in daily life. With this regard, we introduce \textbf{\benchname}, a benchmark for companion-style question answering (QA) on egocentric streaming video. \benchname\ contains 21.6K questions linked to 6K highlight moments across 1,012 long, egocentric videos. It features two key characteristics that are common in daily first-person QA assistance but are largely overlooked in existing VideoQA benchmarks: ego-deictic expressions and interactively chained questions (\eg, “Where is it?”, “How to get there?”). These require models to 
infer a user's in-situation intent by resolving visual pronouns in the context of egocentric visual and QA contents, with both grounded in a long-form streaming setting. 
To tackle the challenges, we propose \textbf{\modelname}, a companion-style QA assistant that highlights a multimodal chain-of-thought reasoning mechanism to infer the final answer based on the historical QA and visual content. An additional question filter and multi-level memory are designed to facilitate efficient QA and visual information retrieval under streaming QA settings. 
Experiments show that \modelname\ significantly enhances the performance of foundation models on \benchname. Moreover, these gains generalize to other streaming and common video QA benchmarks, demonstrating the applicability and effectiveness of our approach. Our code and dataset are available at \url{https://github.com/QHUni/BuddyVQA}.
\end{abstract}

%% file: intro.tex
\section{Introduction}
\label{sec:intro}
\begin{figure*}[t]
\centering
\includegraphics[width=1.02\textwidth]{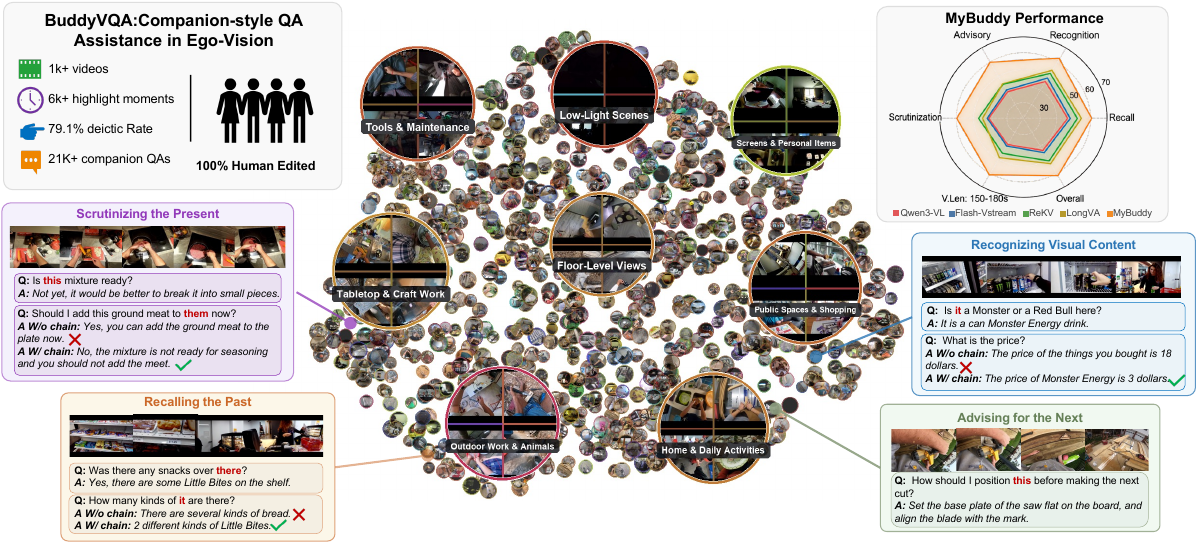}
\vspace{0.3cm}
\caption{\benchname\ targets companion-style VQA assistance. It features two key challenges: ego-deictic expressions and chained questions, both under the egocentric streaming setting. \modelname\ achieves SOTA results across different question categories.
}
\vspace{-0.3cm}
\label{fig:teaser}
\end{figure*}

With the rapid adoption of wearable cameras and mixed-reality devices, egocentric visual assistants have become an important research area for enhancing daily human experiences~\citep{grauman2022ego4d,yang2025egolife,yan2025teleego}. Imagine a visual assistant that accompanies users, continuously observes and remembers their visual experiences, and answers questions when needed (see Fig.~\ref{fig:teaser}).
Often, such assistance requires more than object recognition~\citep{barmann2022did,sun2025visual} and scene description~\citep{xiao2025egoblind,zhou2025egotextvqa}. It requires a companion that maintains visual context over time, understands vague user intentions, and provides practical answers and advice through real-world interaction, serving as an assistive “buddy.”

With this regard, we explore a new setting for companion-style question answering (QA), where the video question answering (VideoQA) model acts as a buddy accompanying the user through egocentric video streams to answer practical questions. 
Such a setting is distinct from traditional VideoQA~\citep{yu2019activitynet,xiao2021next,fu2024video,mangalam2024egoschema}, which focuses on offline, third-person clips with fixed temporal scopes and questions. Recent streaming VQA~\citep{zhang2024flash,lin2024streamingbench,niu2025ovo} is closer in aim, but it ignores egocentric references and conversational continuity in daily QA practice. In contrast, companion-style QA occurs in a shared first-person context, where users frequently ask questions that depend on previously observed events and prior QA interactions. Such questions often contain deictic or under-specified references (\eg, ``\textit{What is this}''), requiring the assistant to infer the user’s true intent.
Therefore, we summarize that a QA buddy should possess 
\textbf{1)} Shared view with egocentric streams;
\textbf{2)} Interpretation of visual deictic expressions;
\textbf{3)} Contextual dependency of questions; and
\textbf{4)} High response efficiency.
Achieving these goals introduces challenges of egocentric referential grounding, multimodal contextual reasoning, and long-term memory, with which
current benchmarks~\citep{mangalam2024egoschema, yang2025egolife, xiao2025egoblind, zhao2025cogstream, yemmego, barmann2022did} do not cover, or cover only partially. 

To bridge this gap, we first construct \textbf{\benchname}, a benchmark targeting egocentric visual QA companions in the streaming setting. \benchname\ contains 21.6K questions linked to 6K moments across 1,012 long ego videos of everyday scenarios.
Each video segment is annotated with temporally aligned questions that emulate how users might query an assistant, from \emph{recall of the past}, to \emph{scrutinization of the present}, to \emph{advising for the future}. 
The benchmark emphasizes two unique phenomena critical to live companion assistance (see Fig.~\ref{fig:teaser}):
\textbf{ego-deictics}, with referential or spatial pronouns (\eg, “this,” “it,” “there”) that must be resolved with respect to the shared egocentric view;
\textbf{interactive chaining}, where successive queries depend on prior QA pairs and visual content, modeling visual conversational continuity. Moreover, the two phenomena are often entangled and grounded in streaming video. 

Benchmarking results show that existing general-purpose and streaming foundation models ~\citep{li2024llava,bai2025qwen2,zhang2024long,li2023videochat,wang2025internvl3,zhang2025videollama} struggle on \benchname; their performance rarely exceeds 50\% accuracy, and the models frequently misinterpret referential expressions or miss important historical information, exposing severe limitations of existing techniques for companion-style assistance. 
For improvements, we introduce \textbf{\modelname}, a training-free and model-agnostic framework designed to tackle the aforementioned challenges for \benchname. 
\modelname\ resolves ego-deictic references following a chain-of-thought (CoT) mechanism that performs visually grounded reasoning in the context of ego-cues (\eg, gaze direction, hand interactions) and relevant historical QAs. Specifically, the ego-cues are extracted from embodied visual features in the memory, which represent the wearer's eye gaze point and hand gestures most relevant to the current question.
The relevant QAs are selected by a Question Filter under the guidance of the current question to support ego-deictic reasoning. Notably, we also propose a multi-level memory to structurally store the visual contents; both the ego-cues and visual contexts are retrieved on demand rather than pre-computed for the entire video stream, enabling efficient online inference under the streaming setting.

Experiments show that \modelname\ achieves state-of-the-art results on \benchname.  It also generalizes well to other streaming and offline VideoQA datasets. The results underscore a significant step towards harnessing foundation multimodal models for live companion-style VQA assistance.
Our primary contributions are:
\textbf{1) \benchname}, a companion-style VQA benchmark that targets egocentric user assistance by highlighting the core challenges of ego-deictics, interactively chained questions, and long memory for streaming QA.
\textbf{2) \modelname}, a training-free and model-agnostic framework for companion-style VQA. It leverages ego-cues for chain-of-thought intention reasoning, driven by multi-level memory to achieve effective and efficient online assistance.
\textbf{3) Key Findings and Improvements}. \benchname\ reveals key limitations in existing models for companion-style VQA, and \modelname\ achieves strong improvements.

%% file: related.tex
\section{Related Works}

\subsection{Streaming VQA}
Streaming VQA aims to continuously interpret incoming video streams and answer questions that may arise at any moment~\citep{zhang2024flash,di2025streaming,xiao2025egoblind,lin2024streamingbench,hu2025streamingcot,niu2025ovo,zhao2025cogstream,yang2026towards}. Unlike traditional offline VideoQA~\citep{xiao2021next,fu2024video,mangalam2024egoschema}, where the entire video is available for question-adaptive processing, Streaming VQA must process videos online without prior knowledge of future questions. Consequently, it relies on a memory buffer to efficiently retain question-agnostic historical video information, enabling the model to answer questions that may be posed at arbitrary future time steps.
Notably, the offline VideoQA task can be treated as a special Streaming VQA case where the questions are posed at the end of the videos. 

To facilitate companion-style QA, \benchname\ follows the streaming VQA setting but differs from existing works by focusing exclusively on egocentric vision (shared user view) and emphasizing a set of cross-modal chaining questions with ego-deictics that reflect the nature of real-life assistance. While the problem of chaining questions has been partially explored in conversational or multi-turn VideoQA~\citep{alamri2019audio, qian2022capturing, pham2022video, christmann2019look, qu2020open}, a key difference and additional challenge in the companion setting is that the cross-modal chained questions and ego-deictics cannot be interpreted from text alone but often require other ego cues from the camera wearer, such as gaze and hand motion information. In addition, both the video streams and questions are assumed infinite, thereby simulating a more practical egocentric QA assistant.

\subsection{Egocentric VQA}
Early egocentric VQA benchmarks primarily evaluate episodic memory, procedural understanding, and long-range reasoning over first-person activities~\citep{grauman2022ego4d, barmann2022did, mangalam2024egoschema}. 
More recent MLLM-oriented studies broaden this scope by collecting egocentric data and evaluating models towards embodied assistance, either semantic ~\citep{cheng2024egothink,zhou2025egotextvqa,yang2025egolife,sun2025visual} or spatial \citep{yang2025thinking}. However, they fail to capture the ego-deictic and chained nature of real-world egocentric question answering.
Although the latest works~\citep{xiao2025egoblind,xiao2026myego,peng2026egogazevqa,choi2026egopointvqa} underscore the importance of gaze and pointing gestures for inferring ego-deictics, these challenges are studied solely through visual information of short video clips. In contrast, our ego-deictics are more practical and complex by naturally entangled with historical QAs rooted in egocentric streaming QA scenarios, which reflects the nature of real-world companion-style QA. Moreover, while existing targeted datasets study ego-gaze or hand pointing independently, \benchname\ reflects both ego-cues in one benchmark.

\subsection{MLLMs for VideoQA}
Recent MLLMs~\citep{an2026llava,bai2025qwen3,qwen3.5,wang2025internvl3, Gemini3.1, GPT-5.5} have substantially advanced offline VideoQA through improved vision-language alignment, temporal encoding, reasoning, and so on. 
Some models further reduce the cost of processing extended inputs through temporal token compression, sparse sampling, hierarchical memory, or retrieval-based reasoning for long video reasoning~\citep{zhang2025sophia,wang2025prolongvid, li2023videochat, zhang2024long, song2025moviechat+, qin2025question, luo2026video}. 
Despite their strong performance, these models favor fixed video reasoning rather than handling dynamically growing video streams. 

Online and streaming Video-LLMs instead process frames incrementally and respond using only the observations up to the question moments. 
Architectures such as Flash-VStream~\citep{zhang2024flash}, Dispider~\citep{qian2025dispider}, and related works~\citep{xiao2023efficient, di2025streaming, kim2025infinipot, xu2025streamingvlm,xiao2026mukv,yeo2026worldmm} introduce mechanisms like memory sinks, dual-process pipelines, and KV-cache reuse for efficient real-time multimodal understanding. Recent corresponding systems~\citep{huang2025online,chatterjee2025streaming,xie2026fluxmem,wang2026streameqa,long2025seeing} try to maintain long-term context under bounded computation by introducing new strategies like compressed memory, frame clustering, and separated perception--decision--response modules. 
While the latest StreamReady~\citep{azad2026streamready} adopts a multi-level memory design, its hierarchy primarily compresses past visual observations into coarse representations to improve storage efficiency, which may discard fine-grained interaction and visual cues critical for assistive video understanding where ego-deictics are frequently observed.
These developments improve online perception and response efficiency, but primarily study single-turn queries, response timing, or memory retention, ignoring ego-deictics and historical QA association.

A related line of work studies video-grounded dialogue and multi-turn multimodal interaction~\citep{qian2022capturing,qu2020open,liu2025taking,lee2025multiverse, sun2026sama,you2025scvbench}. Most such methods condition each response on a concatenation or re-encoding of the preceding dialogue, which becomes increasingly expensive for continuous video streams. More importantly, dialogue history is typically treated as undifferentiated textual context rather than as structured dependencies between current and previous questions. Though recent work investigates prompting strategies with gaze annotations and hand-object detections~\citep{peng2026egogazevqa,xu2024egocentric,inadumi2024gaze}, their focus is mostly on improving detection and localization accuracy, rather than understanding deictic questions with visual entities to support reasoning.
Consequently, \benchname\ presents new challenges of interactive chaining and ego-deictics for real-world QA companions. 
Accordingly, we design a training-free framework \modelname\ tailored for challenges that existing models fundamentally overlook: handling chained question dependencies and ego-deictic expressions in streaming visual context, while maintaining high response efficiency.

%% file: dataset.tex
\section{Dataset: \benchname}
\begin{figure*}[t!]
    \centering
    \includegraphics[width=\textwidth]{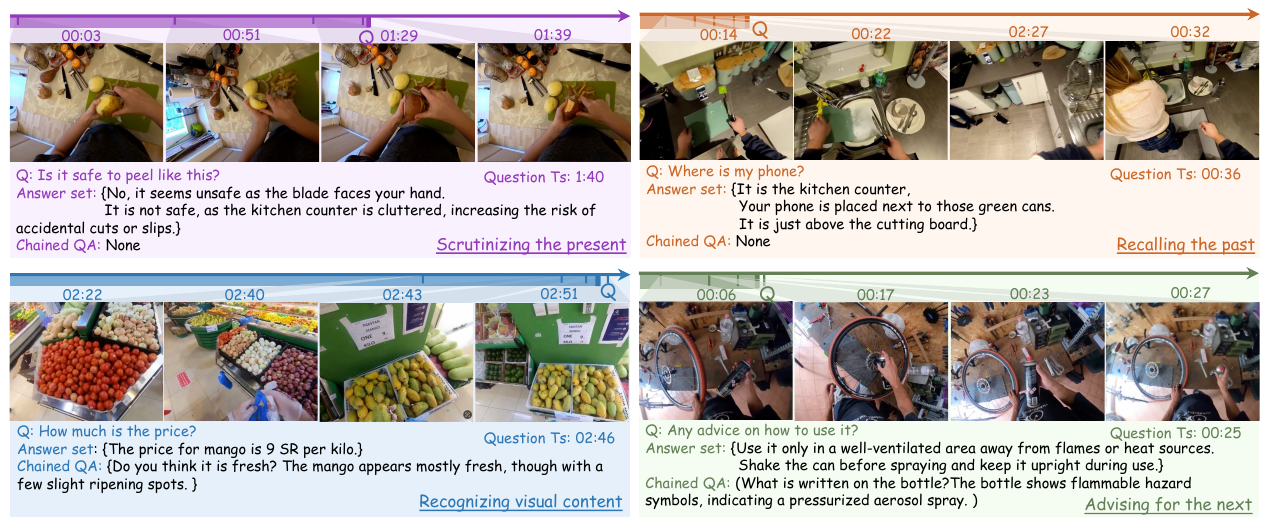}
    \vspace{-0.3cm}
    \caption{Examples of different QAs (Scrutinization, Recall, Recognition \& Advisory) in \benchname.}
    \label{fig:example}
\end{figure*}
To support the study on egocentric companion-style QA, we first construct \benchname\ using videos from EgoSchema~\citep{mangalam2024egoschema} that capture daily egocentric human activities. The QAs are collected to preserve the streaming VQA setting and additionally highlight two core challenges encountered in egocentric companion-style QA:
1) \textbf{Ego-Deictics}, which are referential or spatial pronouns in the egocentric setting, often require visual anchoring of the camera-wearer's QA and activities to resolve. For example, \emph{“What is next to him?”} or \emph{“Is this on the table?”}. This introduces a challenge for robust referential understanding in egocentric QA. 
2) \textbf{Interactively-Chained questions}, which reflect contextual dependencies. The chained questions reference a previous QA pair within the same highlight moment and conversational context. 
This allows for realistic multi-turn dependencies while maintaining temporal coherence.

Prior studies ~\citep{tanenhaus1995integration, clark1996using, tomasello2007shared} in psychology have shown that humans frequently rely on such deictic references when communicating about objects and events in a shared environment. 
In \benchname, the assistant is expected to participate in this shared perceptual space, requiring it to infer the intended referent from the user's egocentric perspective and embodied cues rather than language alone. Consequently, resolving ego-deictic expressions demands not only conversational understanding but also visual grounding and perspective-aware reasoning, all under streaming VQA.

\subsection{Dataset Construction}
\textbf{Stage 1: Video Selection and Automatic Question Generation.} To construct a comprehensive dataset, we select more than 1,000 videos that contain various activities (please refer to the Appendix for detailed activities) and life scenarios from the EgoSchema Dataset~\citep{mangalam2024egoschema}.
To identify key moments in each video, we first employ Gemini-2.5-pro~\citep{comanici2025gemini} to detect visual highlight segments, typically ranging from 2–18 seconds. These highlight moments correspond to perceptually salient or semantically rich events (\eg, object manipulation, transitions, interactions). Fig.~\ref{fig:data-stat}(c) illustrates the distribution of detected highlight moment durations. For each highlight moment, Gemini-2.5-Pro automatically generates multiple question drafts referencing the visual and temporal context. These questions serve as an initial reference for reducing human cognition burden in subsequent annotation. Please refer to the Appendix for the detailed prompts.

\textbf{Stage 2: Human Question Annotation.}
We recruit 25 English-speaking annotators from diverse linguistic and social backgrounds to annotate questions based on the automatically generated QA pairs. Each annotator receives original videos, highlight clips, and associated QA drafts, and follows a standardized procedure: 
a) watch the video segment up to the \textit{end timestamp} of the highlighted moment, without access to future frames. 
b) create questions according to the given QA drafts and video contents, following the principles of
1) \textit{Chained Questions:} If a QA pair is labeled as ``chained'', ensure that this QA pair is closely related to its previously chained QA (not necessarily the immediately preceding one);
2) \textit{Deictic Expressions:}  Adopt conversational phrasing (such as pronouns and deictics) to simulate spontaneous speech; 
3) \textit{Assistive Questions:} Questions should resemble those that a real-world egocentric assistant might encounter;
4) \textit{Diverse Aspects:} Questions within the same clip should address distinct aspects of the event, avoiding simple paraphrases.

The annotations undergo at least two rounds of review and revision, and every accepted QA pair is approved by at least one annotator and one reviewer. In the first annotation round, human annotators produced more than 67K candidate questions. Following a comprehensive expert review, approximately 65\% of these annotations were retained, while the remaining questions were discarded due to issues such as ambiguity, redundancy, or inconsistency with the video content. The retained questions, together with detailed revision comments, were then returned to the original annotators for refinement. After the second annotation round, approximately 29K revised questions satisfied our annotation guidelines. Finally, we performed a quality control stage to further remove semantically redundant questions, resolve duplicate or highly similar QA pairs, and ensure annotation consistency across videos, resulting in 21.6K high-quality qualified questions.

We also ask the annotators to label each question into one of four types:
\textbf{Recall} of past video and QA content,
\textbf{Recognition} of current visual content like objects, text, or people,
\textbf{Scrutinization} of current actions or outcomes, and \textbf{Advisory} feedback for future actions. 
These categories comprehensively cover the past, current, and future time points.

\textbf{Stage 3: Multi-Answer Annotation.}
For the questions reflecting subjective intent (\textit{\eg}, “What should I do next?” or “Can you give me advice on how to cook this?”) in our dataset (common in Scrutinization and Advisory categories), we collect multiple correct answers for each of them, capturing the diversity of plausible responses. Specifically, each question is answered by at least three annotators. Annotators are instructed to use natural spoken language that is consistent with a human-assistant interaction and to avoid leaking information from future video content after the timestamp.
After that, we also conduct a similarity check for all the annotated answers of each question to filter redundant answers. Specifically, we use roberta-large-mnli ~\citep{liu2019roberta} to calculate the pairwise similarity among all the labeled answers to each question, and remove one answer when the pairwise similarity exceeds 0.75.
By applying this approach, we aim to mitigate subjectivity while preserving natural variation in human responses.

\begin{table*}[t!]
\centering
\caption{VideoQA datasets. Chain: chained question. Dei: question containing ego-deictic expressions. EgoV: egocentric videos. MA: multiple ground-truth answers for each question.  QC: question category. OE/MC: Open-Ended/Multi-Choice.}
\label{tab:dataset}
\vspace{0cm}
\resizebox{\linewidth}{!}{%
\begin{tabular}{l c c c c c c c c c}
\toprule
\textbf{Benchmark} & \textbf{V Num.} & \textbf{Total V.} & \textbf{Q Num.}  & \textbf{Dei(\%)} & \textbf{Chain} & \textbf{EgoV} & \textbf{MA} & \textbf{QC} & \textbf{Task} \\
\hline
\multicolumn{10}{l}{\cellcolor{lightpurple!25}\textit{Egocentric VQA Benchmarks}} \\
AssistQ~\citep{wong2022assistq} & 100 & 3.2h & 531 &  8 & \xmark & \cmark & \xmark & \xmark & MC \\
EgoSchema~\citep{mangalam2024egoschema} & 5k & 250h & 5k &  3  &\xmark & \cmark & \xmark & \xmark & MC \\ 
QAEgo4D \citep{barmann2022did} & 1.3k & 180h & 14k  & 5  & \xmark  & \cmark & \xmark  & \xmark & OE \\ 
EgoMemoria \citep{yemmego} & 629 & 149h & 7k &  8 & \xmark  & \cmark & \xmark  & \cmark & MC \\ 
EgolifeQA \citep{yang2025egolife} & 6 & 266h & 6k &  11&  \xmark  & \cmark & \xmark  & \xmark & MC \\ 
EgoGazeVQA~\citep{peng2026egogazevqa} & 913 & - & 1.8k &  7 & \xmark & \cmark & \xmark & \cmark &  MC \\
EgoPointVQA~\citep{choi2026egopointvqa} & 300 & 0.6h & 672 & 25 & \xmark & \cmark & \xmark & \cmark & MC \\
\hline
\multicolumn{10}{l}{\cellcolor{lightblue!25}\textit{Streaming VQA Benchmarks}} \\
VStream-QA~\citep{zhang2024flash} & 32 & 21h & 3.5K &  9 & \xmark  & \xmark & \xmark  & \xmark & OE  \\ 
CogStream~\citep{zhao2025cogstream}  & 1k & 58h & 58k & 4 & \xmark & \xmark  & \xmark  & \cmark &  OE \\ 
StreamingBench~\citep{lin2024streamingbench}  & 900 & - & 4.5k &  2 & \xmark  & \xmark  & \xmark  & \cmark &  MC \\ 
OVO-Bench~\citep{niu2025ovo}  & 644 & 76h & 2.8k  & 4 & \xmark & \xmark  & \xmark  & \cmark &  MC \\ 
EgoBlind~\citep{xiao2025egoblind}  & 1.3k & 15h & 5.3k & 23& \xmark   & \cmark & \cmark & \cmark & OE \\ 
\hline
{\cellcolor{lightorange!25}\textbf{\benchname}} & {\cellcolor{lightorange!25}1k} & {\cellcolor{lightorange!25}50h} & {\cellcolor{lightorange!25}21k} & {\cellcolor{lightorange!25}87.6}  & {\cellcolor{lightorange!25}\cmark}&  {\cellcolor{lightorange!25}\cmark} & {\cellcolor{lightorange!25}\cmark} & {\cellcolor{lightorange!25}\cmark} &  {\cellcolor{lightorange!25}OE} \\
\bottomrule
\end{tabular}
}
 \vspace{-0.3cm}
\end{table*}

\subsection{Dataset Analysis}
\textbf{Dataset statistics.}
\benchname\ is constructed from 1,012 videos selected from EgoSchema~\citep{mangalam2024egoschema}, with 704 videos as the test set and 308 videos as the validation set. It contains a total of 21,574 questions based on 6,002 video highlight moments evenly distributed in the video. The average video duration is 180 seconds, while the average highlight moment duration is 7.48 seconds with 3.6 QA pairs per moment. The average length of question/answer is 8.21/10.62 words, respectively. Each question is associated with an average of 2.76 answers. Our \benchname\ covers a variety of activities from indoor(\eg, cooking, cleaning, laundry, housekeeping) to outdoor(\eg, driving, shopping, gardening, picnics), covering the vast majority of situations encountered in daily assistive contexts. We also found that many vague references are tied to the way people communicate indirectly, either through pointing or eye gaze direction (see Fig.~\ref{fig:example}).

\begin{figure}[t!]
    \vspace{0.2cm}
    \centering
    \includegraphics[width=1.05\linewidth]{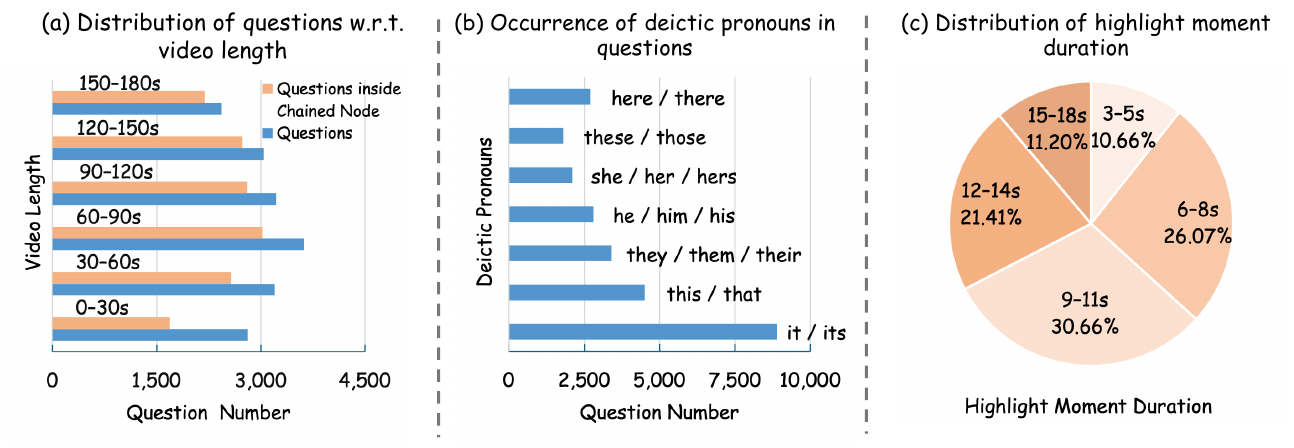}
    \vspace{-0.3cm}
    \caption{Video and question distributions in \benchname.}
    \label{fig:data-stat}
    \vspace{-0.3cm}
\end{figure}

A notable 87.6\% of questions contain at least one deictic pronoun (Fig.~\ref{fig:data-stat}(b)), highlighting the prevalence of implicit references in egocentric communication. 79.1\% of questions are part of a chained node (Fig.~\ref{fig:data-stat}(a)). 
Meanwhile, we also conduct a human study on the validation set to ask humans to answer the chained questions with and without base QAs. The results reveal that 57.6\% of Interactively Chained questions cannot be correctly answered without base QAs, underscoring their dependency. 

Fig.~\ref{fig:data-stat}(a) also shows that the proportion of interactively chained questions increases with video length, reflecting growing contextual dependency over time. Fig.~\ref{fig:example} presents the distribution of question types, indicating a balanced composition and highlighting that \benchname\  contains a significant portion (50.4\%) of subjective (Scrutinization and Advisory) questions. 
At the same time, the majority of chained questions are separated by short to medium gaps, with 37.8\% occurring after 3–5 intervening QAs and 19.3\% after 6–8 intervening QAs, indicating that dependencies are often distributed across multiple turns rather than immediately adjacent. Only 12.7\% are directly consecutive (0 gap), suggesting that modeling longer-range interactions is necessary. Please refer to the Appendix for more detailed analysis.

\textbf{Dataset Comparison.}
~Tab.~\ref{tab:dataset} highlights the key innovation of \benchname: Interactively Chained questions and greater diversity of Ego-Deictics that reflect practical human speech patterns in assistive scenarios. Additionally, \benchname\ provides multiple candidate answers to reduce subjective evaluation, capturing the inherent uncertainty of natural human questioning. These design choices make \benchname\ a more realistic and challenging benchmark for developing companion-based QA systems that must reason dynamically and handle subjectivity in practical scenarios.
In contrast, most existing datasets have a negligible proportion of questions containing ego-deictic expressions. While EgoBlind~\citep{xiao2025egoblind} and EgoPointVQA~\citep{choi2026egopointvqa} exhibit relatively higher proportions, they focus on isolated, history-free visual references without chained cases. Furthermore, EgoBlind is limited to visually impaired users, and EgoPointVQA's questions are restricted to point-based reference queries with mostly synthetic videos. Therefore, both are less representative of the diverse, assistive interactions required in in-the-wild egocentric companion scenarios.

%% file: method.tex
\section{Method: \modelname}
\label{sec:modeling}
\begin{figure*}[t!]
    \centering
    \vspace{0.3cm}
    \includegraphics[width=1\linewidth, height=0.4\linewidth]{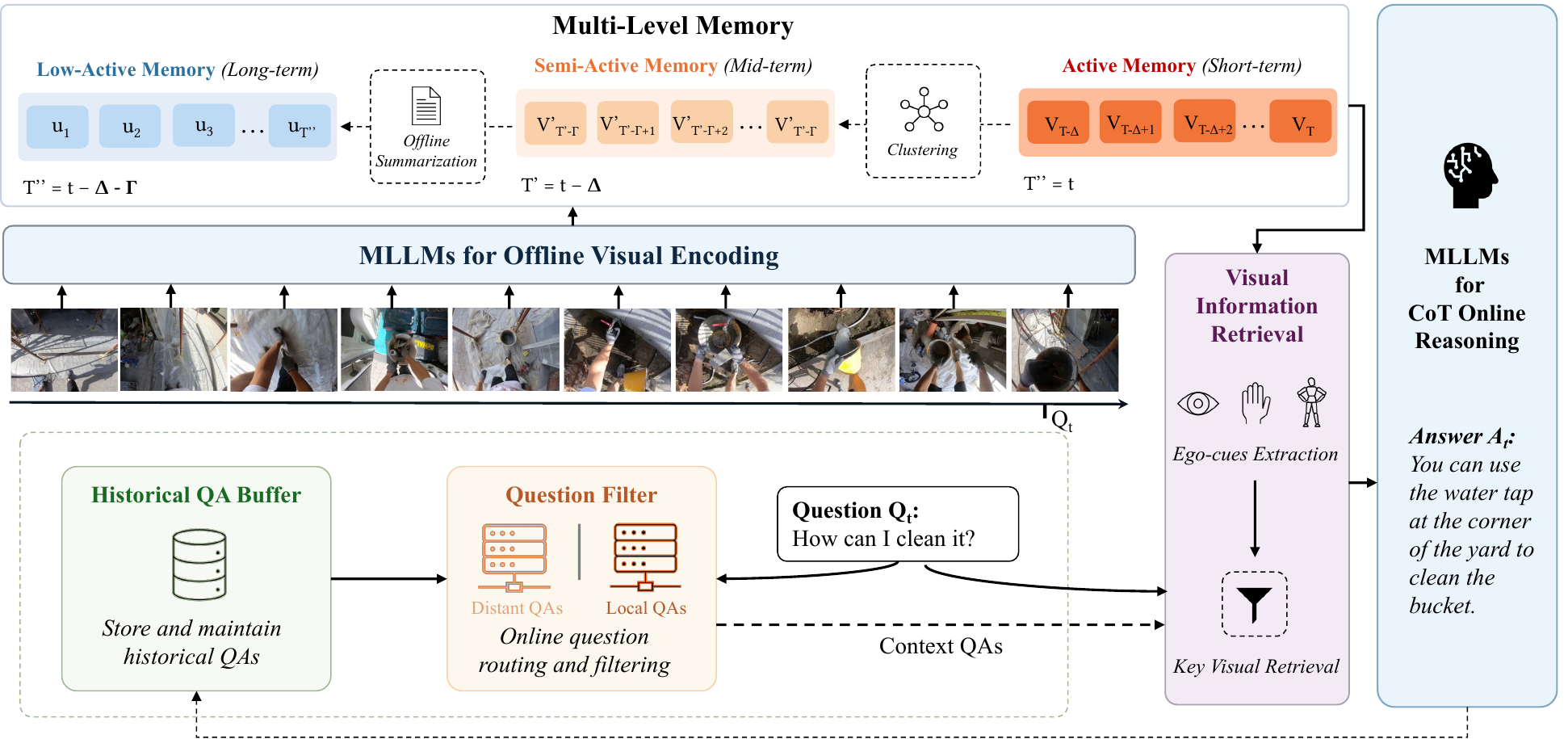}
    \caption{Framework of \modelname. MyBuddy follows a memory-and-retrieval augmented QA pipeline. When a question arrives, it searches from the multi-level memory and historical QA buffer for related visual and textual contexts to infer the question meaning and drive the answer via multimodal chain-of-thought reasoning. } 
    \label{fig:framework}
    \vspace{-0.3cm}
\end{figure*}

\subsection{Overview}
Under the formulation of companion-style QA in streaming setting, we emphasize the following fundamental challenges:
1) \textbf{Deictic and Vague Reference Resolution:} The model needs to identify, through visual cues, how to resolve deictic references.
2) \textbf{Chained Questions:} Reasoning often requires coherent links across historical QAs and visual contexts.
3) \textbf{Response Efficiency:} The model must process and recall streaming sequences without excessive latency or memory expansion when video length grows.

To tackle these challenges, we design a companion-based QA framework (Fig.~\ref{fig:framework}) \modelname. \modelname\ generally follows a memory-and-retrieval augmented QA pipeline and
targets the challenges of \benchname\ with the following additions:
1) \textit{Multimodal CoT Reasoning}, which resolves ego-deictic expressions and rephrases an original vague question into a clear one by associating ego-gaze and hand-pointing regions with relevant historical interaction information.
2) \textit{Question Filter}, which helps identify relevant historical questions to understand a chained question efficiently via a small LLM. 
3) \textit{Multi-Level Heterogeneous Memory}, which improves streaming efficiency (both memory and online QA) by allowing quick and adaptive access to the recent, mid-term, and long-term visual contents in heterogeneous representations (\eg, visual embeddings, dense captions, and summaries). 
Together, \modelname\  maintains a long-term and efficient understanding of the user’s visual experiences to infer their real-time intents, effectively supporting companion-style QA.

\subsection{Multi-Level Heterogeneous Memory}
Since the memory module acts as a prerequisite for supporting contexts to facilitate online QA, we introduce it first for better understanding. Given a sequence of video frames \(  V_t = \{f_1,...,f_t\} \) sampled at a fixed frame rate from a video stream, we obtain their feature representations by a vision encoder
$
    \mathbf{v}_t = \mathcal{E}_v(f_t) \in \mathbb{R}^d
    \label{eq:video_encoder},
$
where \( d \) is the embedding dimension.  
To efficiently store information and support online streaming QA, we design a three-level heterogeneous memory system consisting of \textit{Active}, \textit{Semi-Active}, and \textit{Low-Active} memories:
\begin{equation}
    \mathcal{M} = \{\mathcal{M}_A, \mathcal{M}_S, \mathcal{M}_L\},
\end{equation}
with each maintaining information at a different temporal granularity and semantic abstraction (illustrated at the top of Fig.~\ref{fig:framework}).
Upon usage, memory tiers are prioritized by temporal recency, ensuring that the most recent observations dominate the reasoning process.

\textbf{Active Memory (Short-term,~0--\( \Delta \)).}
The Active Memory (ActMem) retains the most recent visual frame embeddings for frequent access:
\begin{equation}
    \mathcal{M}_A^t = \{\mathbf{v}_{\hat{t}} \mid \hat{t} \in [t-\Delta, t]\},
    \label{eq:active_memory}
\end{equation}
\noindent where \(\Delta\) is a hyperparameter that specifies the temporal span covered by the short-term active memory.

\textbf{Semi-Active Memory (Mid-term,~\(\Delta\)--\( \Gamma\)).}
The Semi-Active Memory (Semi-ActMem) is less visited, and stores aggregated visual contents that exceed the coverage of ActMem for efficiency. 
Regarding aggregation, an incremental version of the incremental density-based clustering algorithm DBSCAN~\citep{ester1998incremental} is applied.  DBSCAN is density-adaptive and allows the model to automatically identify meaningful temporal clusters without predefined thresholds or cluster counts, allowing the number of clusters to grow dynamically along the video stream. Conceptually, Semi-ActMem is obtained by
\begin{equation}
\begin{split}
    & \mathcal{M}_S^t = \text{DBSCAN}(\{\mathcal{M}_S^\tau,\mathcal{M}_A^{\hat{t}}\mid \tau \in[0,t-\Delta], \\ & \hat{t} \in [t-\Gamma-\Delta, t-\Delta]\} \mid \epsilon, minPts),
    \label{eq:dbscan}
\end{split}
\end{equation}
\noindent where $\mathcal{M}_S^{\tau}$ is the maintained Semi-ActMem up to time $\tau$, and $\mathcal{M}_A^{\hat{t}}$ denotes the newly arrived ActMem within the temporal interval $\hat{t}$. Instead of re-clustering all historical memories, we use DBSCAN to incrementally update the maintained clustering state by incorporating the newly arriving memories into the existing density-based clusters. Here, $\epsilon$ denotes the neighborhood radius, while $\mathit{minPts}$ specifies the minimum number of neighboring samples required to form a dense cluster.
Therefore, the semi-active memory at time $\tau$ is represented as a set of density-based clusters 
$\mathcal{M}_S^\tau=\{\mathcal{C}_1^\tau,\ldots,\mathcal{C}_{K^\tau}^\tau\}$.
Each cluster $\mathcal{C}_i^\tau=(\mathbf{r}_i^\tau,\mathcal{E}_i^\tau)$ consists of a representative feature $\mathbf{r}_i^\tau=\frac{1}{|\mathcal{E}_i^\tau|}\sum_{\mathbf{e}\in\mathcal{E}_i^\tau}\mathbf{e},$ and its associated member features $\mathcal{E}_i^\tau$. The representative feature serves as a compact semantic index for efficient retrieval, while the member features preserve fine-grained visual evidence for downstream reasoning.

\textbf{Low-Active Memory (Long-term,~$\geq$$\Gamma$).}
The Low-Active Memory (Low-ActMem) accumulates long-term knowledge through temporally overlapping summaries, allowing recurring activities and persistent contextual information to be naturally reinforced over time.
Specifically, \modelname\ utilizes a zero-shot summarization module \( \mathcal{S} \)  to produce textual summaries and gather visual clues $\mathbf{u}_j$ into memory $\mathcal{M}_L=\{\mathbf{u}_j\}_{j=1}^K$, where
\begin{equation}
     \mathbf{u}_j = \mathcal{S}(\{\mathcal{M}_S^{\hat{t}}\mid \hat{t} \in [t-\epsilon-\Gamma-\Delta, t-\Gamma-\Delta]\}).
    \label{eq:summarization}
\end{equation}
Adjacent summarization windows overlap by $\epsilon$, allowing activities that persist across multiple windows to be repeatedly incorporated into successive summaries and naturally retained in the long-term memory.
\(K\) is the number of summarization windows over the interval from \(T=0\) to \( T=t-\Gamma-\Delta\). Summaries inside the Low-ActMem retain key objects and actions, spatial relations, temporal anchors, and so on to facilitate long-term inquiries (please refer to the Appendix for detailed prompts). 
Unlike ActMem and Semi-ActMem, the summarization process in Low-ActMem is executed during idle intervals (when the QA system is not actively processing incoming questions). Consequently, it functions outside the question–answering pipeline, allowing the model to consolidate contextual information or update memory representations without introducing additional latency to the real-time interaction.

The design of multi-level heterogeneous memory focuses on what to save into memory (features, events, summaries) and how memory is structured (active, semi-active, low-active). 
The three levels of memory effectively retain critical visual representations while simultaneously contributing to improved memory and time efficiency. 

Considering the balance between response efficiency and performance, our design decouples memory maintenance from online question answering. Specifically, the captioning and consolidation of Low-ActMem and the incremental clustering of Semi-ActMem are performed asynchronously during idle intervals, rather than being synchronized with incoming user questions, and therefore introduce no additional response latency. At query time, vision-to-text conversion is restricted to recent ActMem features and question-relevant features selected through the representative cluster centers in Semi-ActMem, while Low-ActMem directly filters its pre-computed textual summaries. Consequently, each query processes only a small set of relevant visual features while retaining multi-scale historical evidence for accurate long-term reasoning. This design limits online computation to question-relevant evidence while preserving both fine-grained recent observations and compact long-term context.

\subsection{Historical QA and Question Filter}
We maintain a single historical QA buffer $B=\{(Q_i,A_i)\}_{i=1}^{t-1}$ to store all QA pairs. For an incoming question $Q_t$, we apply a question filter (\eg, a small MLLM) to identify its relevant QAs from the buffer to serve as contexts. 

Specifically, we dynamically organize the historical QA pairs into \emph{short-term} and \emph{long-term} retrieval candidates according to temporal recency to $Q_t$, thereby characterizing the strength of conversational dependency. We define the distance between a historical QA pair $(Q_i,A_i)$ and $Q_t$ as the number of intervening questions:
\begin{equation}
    d_q(i,t)=t-i-1.
\end{equation}
Then, the query-dependent short-term and long-term QA candidate sets are constructed as
\begin{equation}
\begin{split}  
    B^{\mathrm{s}}(Q_t)
    =\left\{(Q_i,A_i)\in B\mid d_q(i,t)\leq\tau_q\right\}, \\
    B^{\mathrm{l}}(Q_t)
    =\left\{(Q_i,A_i)\in B\mid d_q(i,t)>\tau_q\right\},
\end{split}
\end{equation}
where $\tau_q$ denotes the number of questions covered by the short-term window. 
Note that $B^{\mathrm{s}}(Q_t)$ and $B^{\mathrm{l}}(Q_t)$ are temporary retrieval candidates of $Q_t$, which will be updated dynamically with regard to the incoming questions. 

After that, the question filter first searches only the short-term candidate set for the incoming question $Q_t$:
\begin{equation}
Q_t^{\mathrm{c,s}}=\operatorname{MLLM}\!\left(Q_t,f_t,B^{\mathrm{s}}(Q_t)\right).
\end{equation}
A return is considered meaningful if the filter identifies a historical dependency and retrieves at least one semantically relevant QA pair with a valid identifier from the current candidate set. Otherwise, the filter searches the long-term candidate set for potential long-range dependencies:
\begin{equation}
    Q_t^{\mathrm{c}}
    =
    \begin{cases}
        Q_t^{\mathrm{c,s}},
        & \operatorname{Meaningful}\!\left(Q_t^{\mathrm{c,s}}\right)=1,\\
        \operatorname{MLLM}\!\left(Q_t,f_t,B^{\mathrm{l}}(Q_t)\right),
        & \operatorname{Meaningful}\!\left(Q_t^{\mathrm{c,s}}\right)=0.
    \end{cases}
\label{eq:active_inactive_qa_retrieval}
\end{equation}
If the two-stage search fails to return any relevant historical QA pair, $Q_t$ is treated as an independent question that is directly sent to the main MLLM for answer generation. Otherwise, $Q_t$ is recognized as chained, and the retrieved $Q_t^{\mathrm{c}}$ is used in the subsequent reasoning step. This query-dependent active-first strategy significantly reduces the historical context, while the long-term fallback preserves access to the complete interaction history for long-range chained questions.

\subsection{Multimodal CoT Reasoning}
The CoT mechanism is facilitated by the aforementioned memory and question filter modules. It plays a pivotal role in decoding the deictic expressions and inferring the user's intent behind the questions. 

\textbf{Ego-cues extraction.}
To resolve ego-deictic expressions, the model must infer the object or region implicitly referred to by the camera wearer. In egocentric videos, embodied ego-cues like eye gaze and hand interactions often naturally reveal the wearer's focus of attention. In particular, eye gaze indicates where the wearer is looking, while hand interactions further constrain the object currently being manipulated or referred to. Motivated by this observation, we explicitly extract these ego-cues from the active memory $\mathcal{M}_A$ by prompting MLLM $\mathcal{D}$:
\begin{equation}
\mathbf{V}_t^{ego}=\mathcal{D}(\mathcal{M}_A,Q_t).
\end{equation}
Here, $\mathbf{V}_t^{ego}$ consists of the detected eye gaze point and hand bounding boxes that are relevant to the current question. These ego-cues explicitly indicate the user's focus of attention and are subsequently used to guide the grounding of ego-deictic references before answer generation. Please refer to the Appendix for detailed prompts.

\textbf{Key visual retrieval.}
To achieve CoT reasoning for an incoming question, \modelname\ retrieves key visual information from the multi-level memory. Specifically, we retrieve and generate visual descriptions on demand according to the current question: 
\begin{equation}
\mathbf{V}_t^{des}
=
\left[
\mathcal{D}\!\left(\mathcal{M}_A^t\right);
\mathcal{D}\!\left(\mathcal{O}_S\!\left(\mathcal{M}_S^t,Q_t\right)\right);
\mathcal{D}\!\left(\mathcal{M}_L^t;Q_t\right)
\right].
\label{eq:visual_retrieval}
\end{equation}
For visual features in ActMem $\mathcal{M}_A^t$, we directly convert them into descriptions via a small MLLM $\mathcal{D}$.
For visual features in Semi-ActMem $\mathcal{M}_S^t$, we first filter the top-k cluster centers closest to the question and gather their member features through a two-step retrieval module $\mathcal{O}_S$. 
Specifically, given the question representation $\mathbf{q}_t$, $\mathcal{O}_S$ first computes $\operatorname{sim}(\mathbf{q}_t,\mathbf{r}_i^t)$ for each of the $K_t$ clusters and selects the top-$K$ cluster indices $\mathcal{I}_t$. It then gathers the member features of the selected clusters as $\widetilde{\mathcal{M}}_S^t=\bigcup_{i\in\mathcal{I}_t}\mathcal{E}_i^t$, where $\mathbf{r}_i^t$  and $\mathcal{E}_i^t$ denote the representative feature and the original member video features of cluster $\mathcal{C}_i^t$. 
The retrieved features in $\widetilde{\mathcal{M}}_S^t$ are then converted into descriptions using  $\mathcal{D}$. 
For visual summarization in Low-ActMem $\mathcal{M}_L^t$, we directly filter and identify the descriptions relevant to $Q_t$ using $\mathcal{D}$. 
Both $\mathcal{D}$ and $\mathcal{O_S}$ are implemented in one lightweight MLLM.

\textbf{CoT Reasoning for Answer.}
With the retrieved ego-cues (eye gaze and hand pose) $\mathbf{V}^{ego}_t$, visual information $\mathbf{V}^{des}_t$, and context QAs $Q^{c}_t$, we leverage CoT prompting (please refer to the Appendix for specific prompts) to clarify deictic references, \ie, rephrase the vague question into a disambiguated variant \( \hat{Q}_t\) (\eg, ``where is it?'' to ``where is the cup?''):
\begin{equation}
    \mathcal{\hat{Q}}_t = \mathcal{R}(Q^*_t,\mathcal{G}_t),
\end{equation}
and then answer the rephrased question: 
\begin{equation}
    A_t = \mathcal{R}(\hat{Q}_t,\mathcal{G}_t),
\end{equation}
where $\mathcal{G}_t = [\mathbf{V}^{des}_t, \mathbf{V}^{ego}_t,Q^{c}_t]$, and the MLLM \(\mathcal{R}\) is the main MLLM for answering. 
Note that the resulting answer \(A_t\) will be stored in the historical QA buffer as context for future question answering. 


%% file: experiment.tex
\section{Experiments}
\begin{table*}[t!]
\centering
\small
\setlength{\tabcolsep}{5pt}
\caption{Performance of MLLMs and Humans on~\benchname. \modelname\ outperforms both general and streaming models across all four assistance types, and shows more stable performance as video stream grows. }
\label{tab:baseline}
\vspace{0cm}
\resizebox{\textwidth}{!}{%
\begin{tabular}{llcccccccccccccc}
\toprule
\multirow{2}{*}{\textbf{Model}}&
\multirow{2}{*}{\textbf{Size}}&
\multirow{2}{*}{\textbf{Sampling}}&
\multicolumn{4}{c}{\textbf{Q. Categories}} &
\multicolumn{3}{c}{\textbf{V. Duration}} &
\multirow{2}{*}{\textbf{Overall}} \\
\cdashline{4-7}
\cdashline{8-10} 
&&& \textbf{Recall} & \textbf{Recognition} & \textbf{Advisory} & \textbf{Scrutinization} & \textbf{0-60s} & \textbf{60-120s} & \textbf{120-180s}\\ 
\midrule
{\cellcolor{lightgreen!25}\textit{Human}} & {\cellcolor{lightgreen!25} -} & {\cellcolor{lightgreen!25} -} & {\cellcolor{lightgreen!25} 92.4}  & {\cellcolor{lightgreen!25} 92.1} & {\cellcolor{lightgreen!25} 91.9} & {\cellcolor{lightgreen!25} 92.0} & {\cellcolor{lightgreen!25} 92.0} & {\cellcolor{lightgreen!25} 92.1} & {\cellcolor{lightgreen!25} 92.3} & {\cellcolor{lightgreen!25} 92.1} \\
\midrule
\multicolumn{11}{l}{\cellcolor{lightpurple!25}\textit{General}} \\
VideoChat2~\citep{li2024mvbench} & 8B & 16 & 40.7 & 42.2 & 33.3 & 31.4 & 47.8 & 38.2 & 33.1 & 38.9 \\
Video-LLaMA3~\citep{zhang2025videollama} & 7B & 1 FPS & 41.7 & 44.1 & 36.4 & 34.5 & 48.5 & 40.1 & 34.7 & 39.2 \\
LongVA~\citep{zhang2024long} & 7B & 32 & 46.5 & 46.1 & 39.4 & 40.9 & 47.6 & 40.1 & 39.1 & 43.0 \\
LLaMA-VID~\citep{li2024llamavid} & 7B & 1 FPS & 45.6 & 46.7 & 39.3 & 40.9 & 51.3 & 44.1 & 38.5 & 43.1 \\
LVAgent~\citep{chen2025lvagent} & - & Adaptive & 47.8 & 48.7 & 40.3 & 41.4 & 53.2 & 46.4 & 39.8 & 44.6 \\
Qwen3.5~\citep{qwen3.5} & 9B & 1 FPS & 48.7 & 49.3 & 43.1 & 43.9 & 55.1 & 47.8 & 40.1 & 47.2 \\
Gemini-3.1-pro~\citep{Gemini3.1} & - & 1 FPS & 50.8 & 51.7 & 43.4 & 44.7 & 53.2 & 45.8 & 43.1 & 48.3 \\
InternVL-3.5~\citep{wang2025internvl3} & 8B & 32 & \underline{54.1} & 51.3 & 44.7 & \underline{46.1} & \underline{58.3} & 46.5 & 40.2 & 48.8 \\
GPT-5.5~\citep{GPT-5.5} & - & 60 & 52.2 & \underline{53.4} & \underline{48.9} & 46.0 & 57.3 & \underline{49.2} & \underline{42.1} & \underline{50.5} \\
\hline
\multicolumn{11}{l}{\cellcolor{lightblue!25}\textit{Streaming}} \\
Cogstream~\citep{zhao2025cogstream} & 7B & 1 FPS & 50.2 & 47.6 & 38.2 & 40.1 & 48.2 & 44.0 & 41.6 & 45.0 \\
Dispider~\citep{qian2025dispider} & 7B & 1 FPS & 46.2 & 45.5 & 37.9 & 39.3 & 45.4 & 40.1 & 39.8 & 43.2 \\
Flash-VStream~\citep{zhang2024flash} & 7B & 1 FPS & 52.2 & 51.9 & 44.6 & 46.7 & \underline{54.3} & 49.3 & 45.2 & 50.8 \\
ReKV(LLaVA-OV)~\citep{di2025streaming} & 7B & 1 FPS & \underline{54.6} & \underline{54.1} & \underline{44.9} & \underline{44.4} & 54.0 & \underline{50.1} & \underline{48.6} & \underline{52.7} \\
\hline
\multicolumn{11}{l}{\cellcolor{lightorange!25}\textbf{\modelname(Ours)}} \\
\modelname(Qwen3.5) & 9B & 1 FPS & 59.9 & 60.8 & 55.8 & 54.2 & 61.8 & 59.2 & 58.3 & 60.7 \\
\modelname(Gemini-3.1-pro) & - & 1 FPS & 64.0 & 63.9 & 60.1 & 59.7 & 65.3 & 63.1 & 61.9 & 63.2 \\ 
\modelname(GPT-5.5) & - & 1 FPS & \textbf{65.1} & \textbf{66.5} & \textbf{63.7} & \textbf{63.0} & \textbf{65.9} & \textbf{64.6} & \textbf{63.8} & \textbf{65.1} \\
\bottomrule
\end{tabular}%
}
\vspace{-0.2cm}
\end{table*}

\subsection{Experiment Setup}
\textbf{Evaluation.}
We use GPT-5~\citep{GPT-5} as the answer judge to get the prediction accuracy (0-100\%, the percentage of correct answers) and confidence score (0-5, with 5 being the highest confidence). Notably, our evaluation prompts are manually refined to align closely with human evaluation results for answer scoring (please refer to the Appendix).
We mainly analyze \modelname\ and existing methods on \benchname. For comparison models, we follow their native video sampling protocols whenever applicable. Models supporting a temporal-rate sampling process the observed video prefix at 1 FPS, while fixed-frame models uniformly sample their prescribed number of frames. LVAgent~\citep{chen2025lvagent} adopts its original adaptive retrieval strategy. All models are restricted to video observations before the corresponding question timestamp to ensure fair and reproducible evaluation. We also extend our experiments to VStream-QA~\citep{zhang2024flash}, Video-MME~\citep{fu2024video} and EgoSchema~\citep{mangalam2024egoschema} to validate \modelname's generalization to general streaming and video QA tasks.

\textbf{Implementation.} We choose three strong MLLMs (GPT-5.5~\citep{GPT-5.5}, Gemini-3.1-pro~\citep{Gemini3.1} and Qwen3.5~\citep{qwen3.5}) as our main MLLMs for online CoT reasoning and QA. 
Qwen3.5-2B~\citep{qwen3.5} is employed to obtain the summarizations in the low-active memory, which is performed offline during idle intervals and therefore does not affect online QA latency.
The lightweight MLLM $\mathcal{D}$ used for visual retrieval and question filtering is Qwen3.5-0.8B~\citep{qwen3.5} for the trade-off between efficiency and question-conditioned visual grounding ability.
\( \Delta \) and \( \Gamma \) in the Multi-Level Memory are set to 30s and 120s for \benchname, respectively. $\epsilon$ and $minPts$ in DBSCAN~\citep{ester1996density} are set to 0.27 and 5. 
We set $\tau_q = 8$ by default on \benchname\ based on preliminary analysis. Other evaluated models followed the streaming QA paradigm, receiving video segments and historical QAs chronologically as input. 
All models were evaluated zero-shot, and all hyperparameters are selected on the validation set through preliminary experiments to balance QA accuracy and online efficiency.

\subsection{Result Analysis}

\subsubsection{Comparison on \benchname } 
\begin{figure*}[t!]
    \centering
    \includegraphics[width=1.02\linewidth]{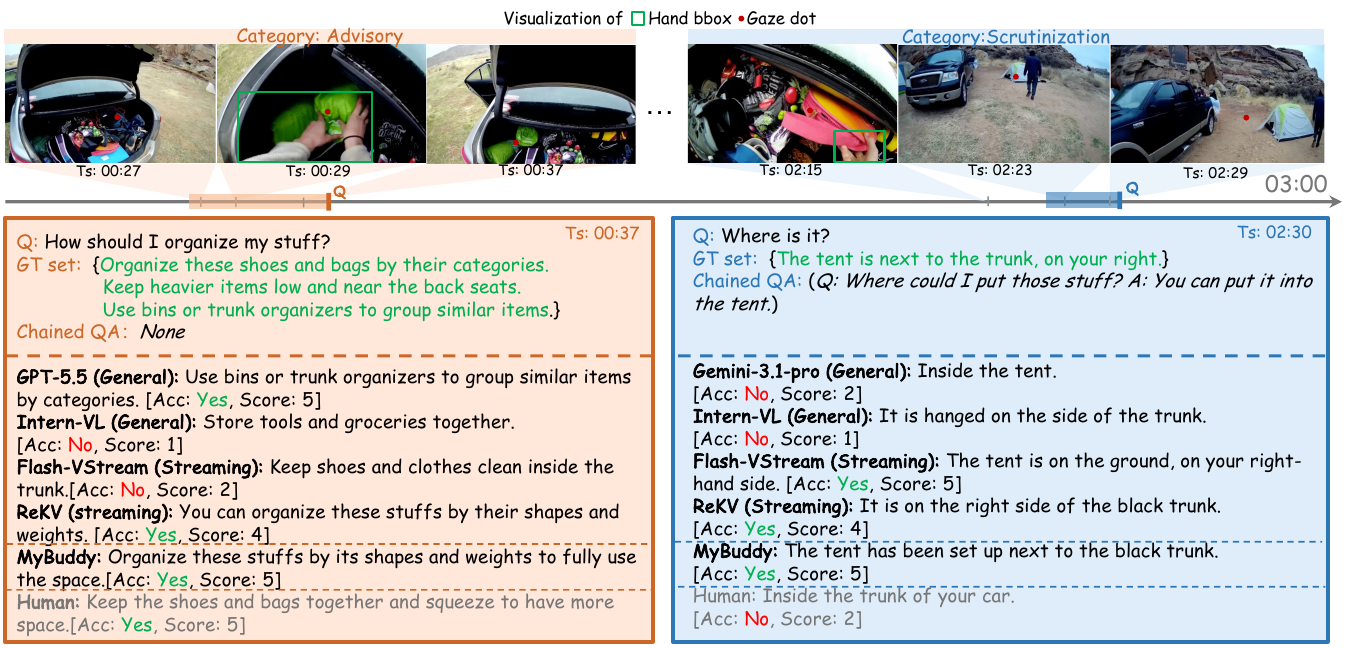}
    \caption{Result visualization on \benchname.  Streaming models generally perform better than general models over time, and \modelname\ is able to process interactively chained questions and ego-deictics regardless of the streaming length. }
    \label{fig:visualization}
    \vspace{-0.2cm}
\end{figure*}

\textbf{Human performance.}
To verify the reliability of \benchname, we first conducted a human evaluation on a random 30\% of questions (as shown in Tab.~\ref{tab:baseline}). Each QA pair was independently reviewed by human evaluators under the streaming setting. The results show a human-verified accuracy of 92.1\%, indicating that most QA pairs align well with visual evidence in each video segment.
Analyses of the remaining 7.9\% of incorrect or ambiguous cases reveal two main error sources:
(a) overly vague ego-deictics amid scenes with multiple similar objects or people, leading to referential confusion (see Fig.~\ref{fig:visualization}); and
(b) motion blur or partial occlusions in some videos cause uncertain visual interpretation.
Despite these imperfections, the strong human–annotation agreement demonstrates that \benchname\ provides reliable and high-quality supervision for the companion-based QA research.

\textbf{Baseline Comparison.} Tab.~\ref{tab:baseline} and Fig.~\ref{fig:visualization} compare \modelname\ with existing general-purpose and streaming VQA models across four question categories and three video duration ranges. 
Overall, \textbf{\modelname\ (GPT-5.5)} achieves the highest accuracy of \textbf{65.1\%}, surpassing the strongest streaming baseline ReKV (LLaVA-OV)~\citep{di2025streaming} by \textbf{12.4\%}. Even other versions—\modelname\ (Qwen3.5) and \modelname\ (Gemini-3.1-pro)—consistently outperform all baselines, reaching 60.7\% and 63.2\%, respectively. Compared to prior streaming models (45.0–52.7\%), \modelname\ achieves substantial gains under the same 1FPS constraint. While general-purpose models like InternVL-3.5~\citep{wang2025internvl3} (48.8\%) and LLAMA-VID~\citep{li2024llamavid} (43.1\%) benefit from larger context windows, they still lag behind, suggesting that frame sampling alone cannot ensure robust temporal reasoning without dynamic context modeling. 

\begin{table}[t!]
\centering
\small
\setlength{\tabcolsep}{1pt}
\caption{Performance comparison on chained and ego-deictic questions in \benchname. 
\modelname\ consistently improves both capabilities, with particularly pronounced gains on ego-deictic questions.}
\label{tab:challenge_results}
\scalebox{0.9}{
\begin{tabular}{lccc}
\toprule
\textbf{Model} & \textbf{Cha. (\%)} & \textbf{Dei. (\%)} & \textbf{All (\%)} \\
\midrule
LongVA~\citep{zhang2024long} & 41.5 & 38.7 & 43.0 \\
VLLaMA3~\citep{zhang2025videollama} & 36.9 & 34.2 & 39.2 \\
LLaMA-VID~\citep{li2024llamavid} & 41.8 & 39.4 & 43.1 \\
Qwen3.5~\citep{qwen3.5} & 43.8 & 42.6 & 47.2 \\
Gemini-3.1-pro~\citep{Gemini3.1} & 46.1 & 42.6 & 48.3 \\
GPT-5.5~\citep{GPT-5.5} & 47.6 & 43.9 & 50.5 \\
InternVL-3.5~\citep{wang2025internvl3} & 47.2 & 43.1 & 48.8 \\
\midrule
Cogstream~\citep{zhao2025cogstream} & 43.8 & 40.2 & 45.0 \\
Dispider~\citep{qian2025dispider} & 41.1 & 38.9 & 43.2 \\
ReKV~\citep{di2025streaming} & 50.9 & 48.6 & 52.7 \\
\midrule
\modelname(Qwen3.5) & 57.3 & 58.9 & 60.7 \\
\modelname(Gemini-3.1-pro) & \underline{59.7} & \underline{60.4} & \underline{63.2} \\
\modelname(GPT-5.5) & \textbf{62.8} & \textbf{63.5} & \textbf{65.1} \\
\bottomrule
\end{tabular}
}
\vspace{-0.15cm}
\end{table}

\begin{table*}[t!]
\setlength{\tabcolsep}{10pt}
\centering
\small
\caption{
Dynamic streaming evaluation on BuddyVQA.
We report accuracy (\%) under three settings for better comparisons:
1) \emph{No Feedback:} No user feedback of historical questions, 
2) \emph{Correctness:} Binary feedback (correct or not) of  historical questions, 
and
3) \emph{GT Answer:} Ground-truth answers of historical questions are provided by users.
The best and second-best results are highlighted in bold and underline, respectively.
}
\label{tab:dynamic_streaming}
\begin{tabular}{l c ccc ccc}
\toprule
\multirow{2}{*}{\textbf{Model}}
& \multicolumn{1}{c}{No Feedback}
& \multicolumn{3}{c}{Correctness}
& \multicolumn{3}{c}{GT Answer} \\
\cmidrule(lr){2-2}
\cmidrule(lr){3-5}
\cmidrule(lr){6-8}
& Overall
& Overall & $0$--$90$s & $>90$s
& Overall & $0$--$90$s & $>90$s \\
\midrule
Qwen3.5~\citep{qwen3.5} & 47.2 & 49.0 & 51.8 & 45.9 & 52.3 & 54.2 & 50.1 \\
Gemini-3.1-pro~\citep{Gemini3.1} & 48.3 & 50.4 & 53.0 & 47.4 & 54.0 & 55.9 & 51.7 \\
GPT-5.5~\citep{GPT-5.5} & 50.5 & 53.1 & 55.0 & 50.8 & 56.9 & 58.2 & 55.3 \\
\midrule
MyBuddy (Qwen3.5) & 60.7 & 63.2 & 63.8 & 62.5 & 67.1 & 66.2 & 68.2 \\
MyBuddy (Gemini-3.1-pro) & \underline{63.2} & \underline{66.0} & \underline{66.4} & \underline{65.5} & \underline{69.8} & \underline{68.8} & \underline{70.9} \\
MyBuddy (GPT-5.5) & \textbf{65.1} & \textbf{68.0} & \textbf{68.3} & \textbf{67.6} & \textbf{72.1} & \textbf{70.9} & \textbf{73.5} \\
\bottomrule
\end{tabular}
\end{table*}
Across question categories, \modelname\ shows balanced improvements. Most existing models struggle with Advisory and Scrutinization types (6.7–9.2\% below Recall and Recognition), reflecting the challenge of subjective reasoning. In contrast, \modelname\ achieves 63.7\% and 63.0\% on these types, outperforming ReKV (LLaVA-OV)~\citep{di2025streaming} by 18.8\% and 18.6\%, respectively. This indicates strong capability in handling subjective, context-dependent queries through its hierarchical prompting and memory-based context integration.
Regarding video duration, most models degrade over time. For example, ReKV (LLaVA-OV)~\citep{di2025streaming} drops from 54.0\% to 48.6\%, and InternVL-3.5~\citep{wang2025internvl3} from 58.3\% to 40.2\% when the video length increases from 60s to 180s. This suggests that the accumulation of visual and contextual information over longer sequences imposes memory and reasoning burdens that most models are not equipped to handle. Their intermediate representations drift, temporal dependencies are not retained reliably, and key details from earlier segments become inaccessible.
In contrast, \modelname\ (GPT-5.5) remains stable (64.6\%→63.8\%), demonstrating superior temporal consistency and long-term dependency modeling. 

We further analyze \modelname\  on the subset of chained and ego-deictic questions. Table~\ref{tab:challenge_results} shows that \modelname\ consistently outperforms all baselines on both challenging subsets. With GPT-5.5, \modelname\ exceeds the strongest baseline ReKV~\citep{di2025streaming} by 11.9\% and 14.9\% on chained and deictic questions, respectively. The gains are especially pronounced on ego-deictic queries, validating the strength of combining ego-cues with history-aware reasoning. Moreover, the improvements remain consistent across different backbones, demonstrating \modelname's model-agnostic effectiveness.

\subsubsection{Dynamic Streaming Evaluation}
To analyze companion QA performance in realistic assistive scenarios, we additionally conduct a human-in-the-loop evaluation and use ground-truth (GT) answers to simulate the human feedback of historical QAs in streaming environments. We analyze and compare three settings: 
1) No feedback (by default): the historical questions and model predictions are used as contexts for answering the current question,
2) Binary feedback:  the historical questions and model responses, together with the correctness of the responses, are used as contexts,
and 3) GT Answer: the historical questions and their corresponding GT answers are used as contexts.
As shown in Table~\ref{tab:dynamic_streaming}, \modelname\ consistently benefits more from user feedback than the standalone backbone models. With \emph{correctness} feedback, \modelname\ (GPT-5.5) improves from 65.1\% to 68.0\%, showing that even a binary signal helps the model identify unreliable historical responses and avoid propagating previous errors. When users directly correct the wrong answers with the correct ones, the accuracy increases to 72.1\%, outperforming GPT-5.5 under the same setting by 15\%. The advantage is most pronounced after 90 seconds: while the standalone backbones still suffer from error accumulation in the growing interaction history, \modelname\ (GPT-5.5) reaches 73.5\%, exceeding its 0--90s performance of 70.9\%. Similar improvements are observed with the Qwen3.5 and Gemini-3.1-pro backbones, demonstrating that the benefit is model-agnostic. We attribute these gains to MyBuddy's question filter and structured multi-level memory, which retain useful corrections, suppress irrelevant or incorrect historical responses, and retrieve the feedback most relevant to subsequent chained questions.

\begin{table*}[t!]
\small
\setlength{\tabcolsep}{11pt}
\centering
    \caption{Comparison on RVS-Ego and RVS-Movie~\citep{zhang2024flash}. Acc.: Accuracy. Sco.: Score. Latency \& GPU Usage are tested on Tesla A100 GPU. Results are averaged on 10 videos with each 720 frames as inputs. }
    \label{tab:RVS}
    \begin{tabular}{l c c c c c c }
        \toprule
        \multirow{2}{*}{\textbf{Model}}&
        \multicolumn{2}{c}{\textbf{RVS-Ego}} &
        \multicolumn{2}{c}{\textbf{RVS-Movie}} &
        \multirow{2}{*}{\textbf{Latency} $(\downarrow)$} &
        \multirow{2}{*}{\textbf{GPU Usage} $(\downarrow)$} \\
        \cdashline{2-3}
        \cdashline{4-5}
        & Acc. & Sco. & Acc. & Sco. \\
        \hline
        Flash-VStream~\citep{zhang2024flash} & 58.9 &4.0 & 56.0 &3.3 & \underline{2.4s} & \underline{20GB} \\
        ReKV(LLaVA-OV)~\citep{di2025streaming} & 63.7&4.0 & 54.4&3.6 & 3.3s & 38GB \\
        LongVA~\citep{zhang2024long} & 53.2& 3.1 & 43.0 &3.2 & 6.3s & 37GB \\
        MovieChat~\citep{song2025moviechat+} & 52.2 &3.4 &39.1 &2.3 & 8.8s & 40GB \\
        LLaMA-VID~\citep{li2024llamavid} & 53.4& 3.9 & 48.6 &3.3 & 11.9s & 33GB \\
        Qwen3.5~\citep{qwen3.5} & 57.2 & 3.3 & 51.6 & 3.3 & 14.8s & 51GB \\
        GPT-5.5~\citep{GPT-5.5} & 59.2&3.8 & 48.2&3.4 & - & - \\
        \hline
        \modelname\ (Qwen3.5) & 64.8&3.7 & 55.8&\textbf{3.7} & 3.0s & 23GB \\
        \modelname\ (Gemini-3.1-pro) & 65.3& 3.7 & 57.2& 3.6 & 3.4s & - \\
        \modelname\ (GPT-5.5) & \textbf{66.1}& \textbf{3.9} & \textbf{58.9}& 3.6 & 3.4s & - \\
        \bottomrule
        \end{tabular}
        
\end{table*}
\subsubsection{Extension to General Streaming and Video QA} 
\textbf{RVS-Ego and RVS-Movie.} To further validate the streaming performance on long videos, we also evaluate \modelname\ on VStream-QA~\citep{zhang2024flash} comprising RVS-Ego and RVS-Movie, where the video length ranges from 30 minutes to 60 minutes. Tab.~\ref{tab:RVS} shows that \modelname\ (GPT-5.5) achieves SOTA accuracy of 66.1\% on RVS-Ego and 58.9\% on RVS-Movie, outperforming all competitors, including base GPT-5.5 \citep{GPT-5.5}. 
While Flash-VStream~\citep{zhang2024flash} attains a 2.4s latency, \modelname\ models operate at 3.0–3.4s, offering a balanced trade-off between speed and accuracy. All these results emphasize the overall abilities of \modelname\  on long streaming videos with different characteristics.

\textbf{Video-MME and EgoSchema.} 
For validating and extending the generalization of \modelname\ beyond companion-based QA, we additionally evaluate its performance on standard VideoQA datasets Video-MME~\citep{fu2024video} and EgoSchema~\citep{mangalam2024egoschema}.
We extend \modelname\ to these datasets by treating the full video as a completed stream before the question. Tab.~\ref{tab:generalization} shows that \modelname\ demonstrates competitive and, in several cases, superior performance compared to existing SOTA methods across multiple benchmarks. On Video-MME, \modelname\ (Gemini-3.1-pro) surpasses strong baselines such as LVAgent~\citep{chen2025lvagent} (81.7\% / 86.6\%) and GPT-5.5~\citep{GPT-5.5} (77.3\% / 82.6\%), while maintaining consistent performance across datasets. On EgoSchema, \modelname\ (GPT-5.5) reaches 82.9\%, matching the best reported performance of LVAgent~\citep{chen2025lvagent} and exceeding other large-scale MLLMs like GPT-5.5~\citep{GPT-5.5} (79.8\%). These results indicate that \modelname\ also performs competitively on general VideoQA benchmarks.

\begin{table}[t!]
    \vspace{-0.4cm}
    \centering
    \caption{Model comparison on Video-MME~\citep{fu2024video} and EgoSchema~\citep{mangalam2024egoschema}. Video-MME: ``without subs / with subs". The best and second-best Accuracy results are \textbf{bolded} and \underline{underlined}, respectively. }
    \resizebox{\linewidth}{!}{%
    \label{tab:generalization}
    \setlength{\tabcolsep}{9pt}
    \begin{tabular}{lcc}
    \toprule
    \textbf{Model} & \textbf{Video-MME} & \textbf{EgoSchema} \\
    \midrule
    Video-LLaMA3~\citep{zhang2025videollama} & 47.9 / 49.7 & 57.3 \\
    VideoChat2~\citep{li2024mvbench} & 39.5 / 43.8 &56.7 \\
    MovieChat~\citep{song2025moviechat+} &  38.2 / - & 53.5 \\
    Qwen3.5~\citep{bai2025qwen2} & 71.2 / 74.3 & 68.1 \\
    GPT-5.5~\citep{GPT-5.5}       & 77.3 / 82.6 & 79.8 \\
    \midrule
    LongVA~\citep{zhang2024long}      & 52.6 / - & - \\
    Video-XL~\citep{shu2025video}   & 55.5 / 61.0 & - \\
    Dispider~\citep{qian2025dispider} & 57.2 / - & 55.6 \\
    LVAgent~\citep{chen2025lvagent}     & \ul{81.7} / \ul{86.6} & \textbf{82.9} \\ 
    \midrule
    \modelname\ (Qwen3.5) & 71.3 / 75.5 & 47.8 \\
    \modelname\ (Gemini-3.1-pro) & \textbf{82.5} / \textbf{87.3} & 82.2 \\
    \modelname\ (GPT-5.5) & 81.1 / 84.9 & \textbf{82.9} \\
    \bottomrule
    \end{tabular}
}
\end{table}

\subsection{Ablation Study}
\label{sec:modeanalysis}
We ablate \modelname\ on \benchname\ \wrt interactively chained questions and questions containing ego-deictic expressions in Tab.~\ref{tab:ablation}. 
The largest performance drop occurs when historical QA is not given, causing accuracy fall from 62.8\% to 44.5\% on chained questions, highlighting the importance of temporal linkage across chained QA pairs. In contrast, unchained questions remain relatively stable with a minor performance drop (-4.0\%), confirming that historical QA specifically benefits chained reasoning.
Meanwhile, removing the hierarchical rephrasing results in noticeable declines for both chained (-10.1\%) and deictics (-12.6\%) questions, demonstrating the necessity of linguistic reformulation for resolving referential queries. Similarly, ablating embodied ego-cue extraction reduces deictic accuracy by 9.1\%, showing the usefulness of embodied cues that enhance grounding of ego-deictics by guiding attention to salient visual regions. The multi-level memory ablations further reveal an efficiency–accuracy trade-off. 
\begin{table}[t!]
    \vspace{-0.35cm}
    \centering
    \caption{Ablation of \modelname\ on \benchname\ w.r.t interactively chained questions and questions with deictic expressions. AM/SM/LM: Active/Semi-Active/Low-Active Memory. For the setting of w/o SM \& LM \& AM, we use the question frame as input. }
    \resizebox{\linewidth}{!}{%
    \label{tab:ablation}
    \small
    \setlength{\tabcolsep}{7pt}
    \begin{tabular}{l c c c c}
        \toprule
        \multirow{2}{*}{\textbf{Ablation Model}}&
        \multicolumn{2}{c}{\textbf{Chain (\%)}} &
        \multicolumn{2}{c}{\textbf{Deictic (\%)}} \\
        \cdashline{2-3}
        \cdashline{4-5}
        & \xmark & \cmark & \xmark & \cmark \\
        \hline
        \textbf{\modelname(Full)}  & \textbf{66.4} & \textbf{62.8} & \textbf{66.1}& \textbf{63.5} \\
        w/o LM & 65.7 & 61.3 & 63.3 & 58.9 \\
        w/o SM \& LM & 64.1 & 58.0 & 57.8 & 55.8 \\
        w/o AM \& SM \& LM  & 59.2 & 53.1 & 53.4 & 52.9 \\
        \hline
        w/o Historical QA & 62.4 & 44.5 & 61.3 & 54.3 \\
        w/o Embodied Ego-cues & 55.3 & 54.2 & 56.2 & 54.4 \\
        w/o Hierarchical Rephrasing & 54.7 & 52.7 & 58.0 & 50.9 \\
        \bottomrule
    \end{tabular}
    }
    \vspace{-0.3cm}
\end{table}

\begin{figure}[t!]
    \centering
    \includegraphics[width=\linewidth]{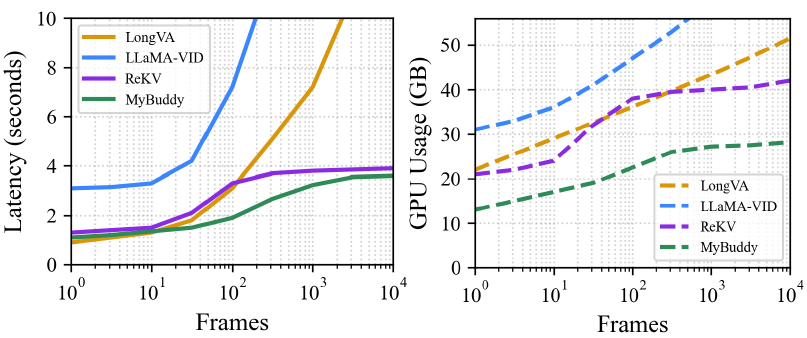}
    \caption{Streaming VQA latency \& GPU Usage \wrt number of frames, tested on A100 GPUs. \modelname\ is able to process long streams more efficiently.}
    \label{fig:efficiency}
\end{figure}

\textbf{Efficiency Analysis.}
While efficiency is vital for the streaming setting, we measure the latency and GPU usage with respect to video length for \modelname\ efficiency analysis, as shown in Fig.~\ref{fig:efficiency}. 
Latency is measured from question input to response completion. GPU Usage indicates peak GPU memory usage during inference.
\modelname\ demonstrates a clear advantage in both latency and GPU efficiency compared to existing streaming and long-context models, with the employment of multi-level memory and an on-demand processing mechanism that are beneficial for practical deployment. For general MLLMs such as LLaMA-VID~\citep{li2024llamavid}, the latency scales up with the number of frames as their structures require processing all frames in a single pass~\citep{zhang2024flash}. Different from them, Streaming VQA models like ReKV~\citep{di2025streaming} enable processing streaming videos of more than 1,000 frames, yet still occupy a considerable amount of computational resources. 

\begin{table*}[t!]
\vspace{-0.2cm}
\small
\centering
\setlength{\tabcolsep}{12pt}
\caption{Efficiency breakdown on A100 GPUs. Extract Cost denotes additional overhead (latency/memory) introduced by individual modules. Base (Qwen3.5-9B): processes video frames \& concatenates all historical QAs for each question online.}
\label{tab:break-efficiency}
\begin{tabular}{lccccc}
\toprule
\textbf{Configuration} & \textbf{Acc.}  & \textbf{Chained Acc.} & \textbf{Latency}  & \textbf{GPU Usage}  & \textbf{Extra Cost} \\
\midrule
Base (Raw F+His. QAs) & 47.2 & 43.8 & 3.9s & 51.1GB & -- \\
+ Multi-Level Memory & 51.4 & 48.2 & 2.5s & 22.9GB & -1.4s / -28.2GB \\
+ Question Filter & 53.6 & 54.5 & 2.3s & 30.2GB & -1.6s / -20.9GB\\
+ Key Visual Info Retrieval & 55.7 & 52.7 & 4.0s & 62.3GB & +0.1s / +11.2GB\\
+ Hierarchical Rephrasing & 56.9 & 53.4 & 4.3s & 53.8GB & +0.4s / +2.7GB \\
\midrule
\textbf{\modelname (Full)} & \textbf{60.7} & \textbf{57.3} & \textbf{2.6s} & \textbf{29.1GB} & \textbf{-1.3s / -12.0GB} \\
\bottomrule
\end{tabular}
\vspace{-1mm}
\end{table*}
To further analyze the latency and computational cost of each module in \modelname, we also conduct a breakdown efficiency analysis (as shown in Tab.~\ref{tab:break-efficiency}). Results show that our module design offsets the additional cost resulting from retrieval and question rephrasing, as they dispense with online long video processing and long-context LLM inference. For example,
Tab.~\ref{tab:break-efficiency} Row 1 (Base) \vs Row 3 shows that the Question Filter brings advantages in efficiency (-1.6s, -20.9GB) and accuracy (+6.4\%, especially chained accuracy, +10.7\%) compared with naive concatenation, resulting from significantly reduced noise and context length. Overall, the performance gains from modules in \modelname\ outweigh the costs in system complexity and inference time.

\subsection{Case Study}

\begin{figure*}[t!]
\centering
\includegraphics[width=\textwidth]{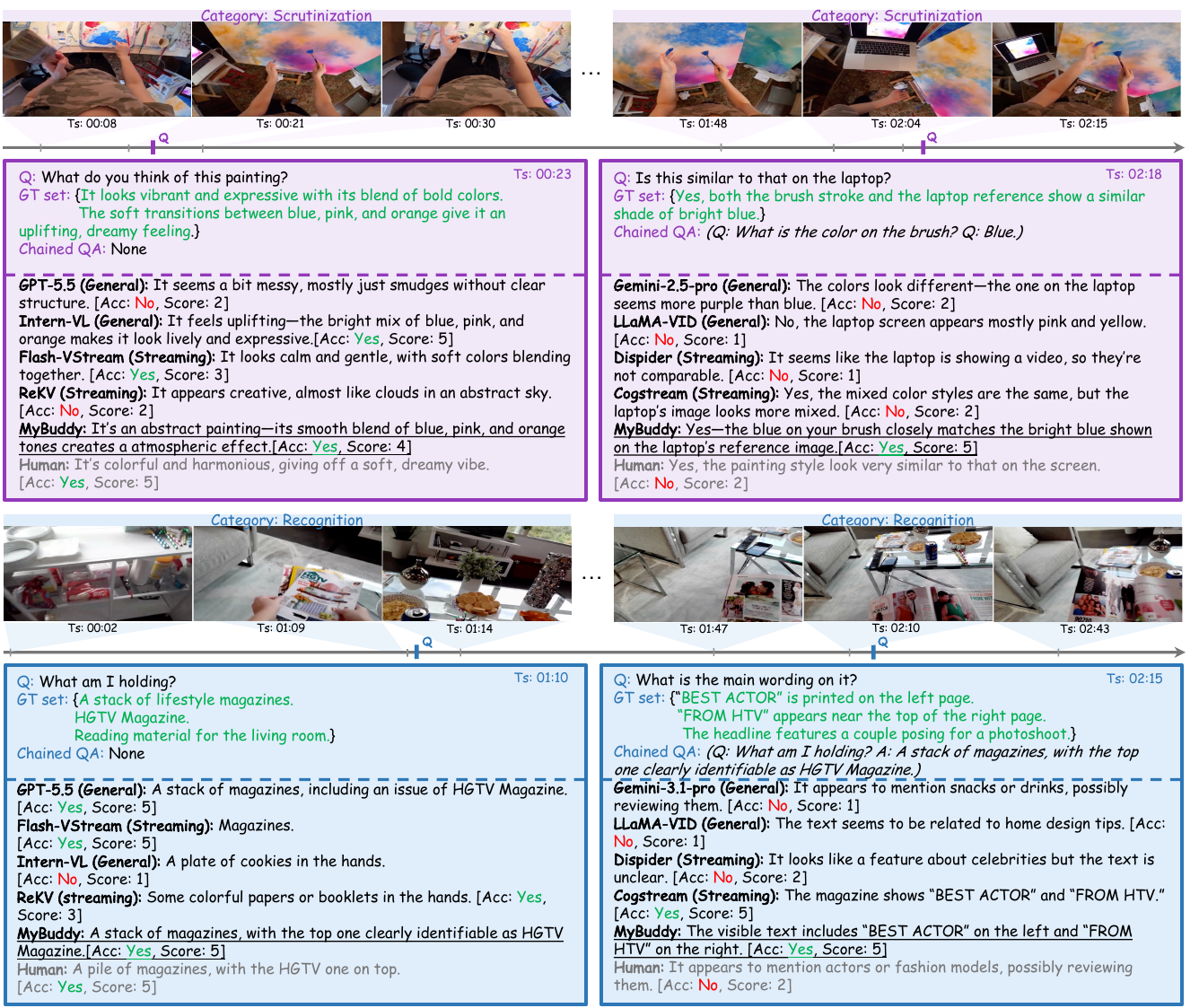}
\caption{More result visualizations on \benchname\ in \textit{Scrutinization} and \textit{Recognition} categories. Baseline models perform poorly on deictic questions, underscoring the importance of jointly leveraging historical interaction contexts and embodied visual cues for ego-deictic reference resolution.}
\label{fig:supp_vis_a}
\vspace{-4mm}
\end{figure*}

\begin{figure*}[t!]
\centering
\includegraphics[width=\textwidth]{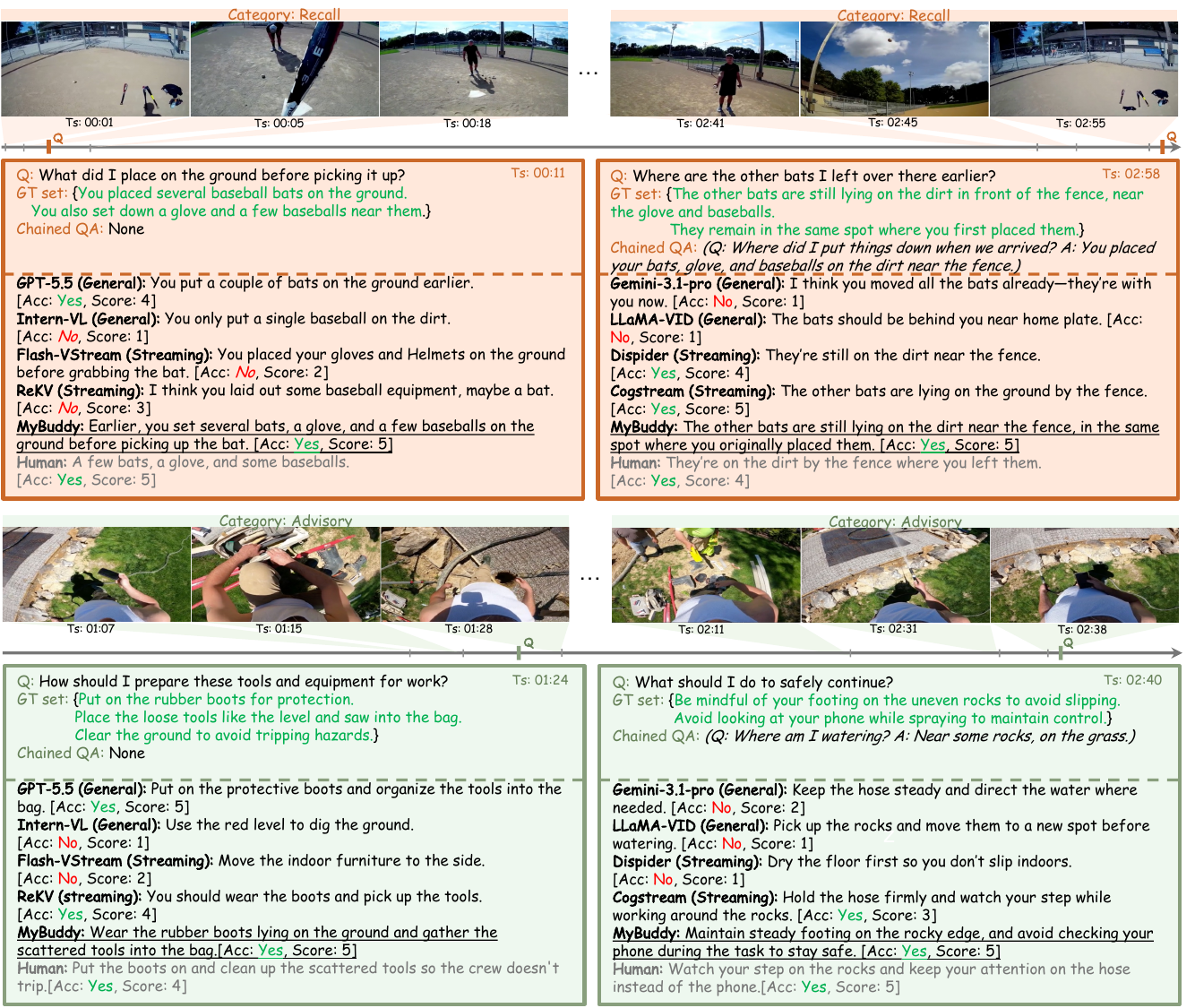}
\caption{More result visualizations on \benchname\ in \textbf{Recall} and \textbf{Advisory} categories. \modelname\ demonstrates substantially stronger performance on long video sequences compared to general MLLMs. Its consistent improvements on chained questions further highlight the importance of history-aware reasoning across question chains in \benchname.}
\label{fig:supp_vis_b}
\vspace{-4mm}
\end{figure*}

\begin{table}[t!] 
\setlength{\columnsep}{6pt}
\centering
\small
\caption{Step-wise failure analysis of \modelname\ on the subset of \benchname. While early-stage errors exist, a significant portion of failures arises from intrinsic ambiguity and visual challenges, rather than pipeline design alone. }
\resizebox{1.0\linewidth}{!}{
\begin{tabular}{lccc}
\toprule
\textbf{Failure} & \textbf{Num} & \textbf{Ratio (\%)} & \textbf{Main Reasons} \\
\midrule
Que. Filter & 124 & 6.2 & Misclassifying chained questions \\
Vis. Retri.  & 420 & 21.0 & Missing relevant frames or objects \\
Rephrase & 504 & 25.2 & Wrong referent grounding \\
Ans. Infer & 556 & 27.8  & Correct context, wrong inference \\
Others & 396 & 19.8  & Motion blur / Noisy input \\
\bottomrule
\end{tabular}
}
\label{tab:failure_analysis}
\end{table}

\textbf{Failure case analysis.}
We first carry out a manual failure analysis over 2,000 \benchname~failure cases from the validation set by examining error outputs at each stage, and trace the causes in Tab.~\ref{tab:failure_analysis}. These results suggest that the Question Filter has the lowest failure rate, indicating limited error propagation from early stages.  
Most failures originate from rephrasing and answer inference, showing that ego-deictic grounding and chained reasoning are the core bottlenecks, which is consistent with the intended diagnostic focus of the \benchname~benchmark. 

\textbf{Case visualizations.}
We also visualize additional qualitative results across the major categories in \benchname~in Fig.~\ref{fig:supp_vis_a} and Fig.~\ref{fig:supp_vis_b}.
In many cases, we observe that deictic expressions do introduce a substantial challenge for existing VideoQA models. Interestingly, even when two questions are semantically identical and differ only in the way the referent is expressed (i.e., an explicit noun versus a deictic expression), most existing MLLMs often exhibit a substantial performance gap. This is also reflected in the differing performance gaps between \modelname\ and baselines on \benchname (\eg, \modelname\ 65.1\% vs GPT-5.5 50.5\%) and on general VideoQA benchmarks (\eg, \modelname\ 84.9\% vs GPT-5.5 82.6\% on Video-MME).
Error analysis indicates that the majority of failures for current models originate from incorrect referent grounding rather than deficiencies in semantic reasoning. Specifically, the models frequently confuse visually similar objects, persons, events, or scene contexts, causing subsequent reasoning to operate on incorrect evidence. These findings demonstrate that deictic expressions expose a fundamental limitation of current VideoQA systems and motivate the need for dedicated methods that jointly resolve referential ambiguity and perform grounded reasoning.

For chained questions, we find that the difficulty is further amplified, as many baseline models struggle to correctly interpret the referential structures embedded in multi-turn dialogue. In scenarios where ego-deictic expressions point back to earlier answers, correct grounding requires explicit reasoning over the past QA context rather than isolated single-step inference. Some baseline models fail to genuinely resolve the intended referent, leading to semantically incorrect responses despite the relevant visual information being present in the video. These errors underscore the importance of robust co-reference understanding and dialogue-aware reasoning in egocentric interactions, where ego-deictic phrasing is pervasive and cannot be resolved without jointly leveraging both historical visual evidence and prior linguistic exchanges.

At the same time, we observe that humans, despite their strong overall performance, can still make mistakes when subjected to challenging visual conditions such as motion blur, rapid camera movement, or the presence of very small or partially occluded text. These fine-grained visual cues are difficult to perceive even for human annotators and highlight the need to develop models robust to the intrinsic ambiguity present in real-world egocentric videos, where frames are not always clean, stable, or fully interpretable. Also, for some extremely abstract deictic expressions in companion QA, humans may also make mistakes.
In such cases, the referent may be very ambiguous or only implicitly grounded in earlier contexts, making the correct interpretation genuinely difficult. These situations highlight that resolving deictic cues is not only a challenge for models but also a non-trivial cognitive task for humans.

\subsection{Evaluation Agreement between Human and LLM}
\begin{figure}[t!]
    \centering
    \includegraphics[width=\linewidth]{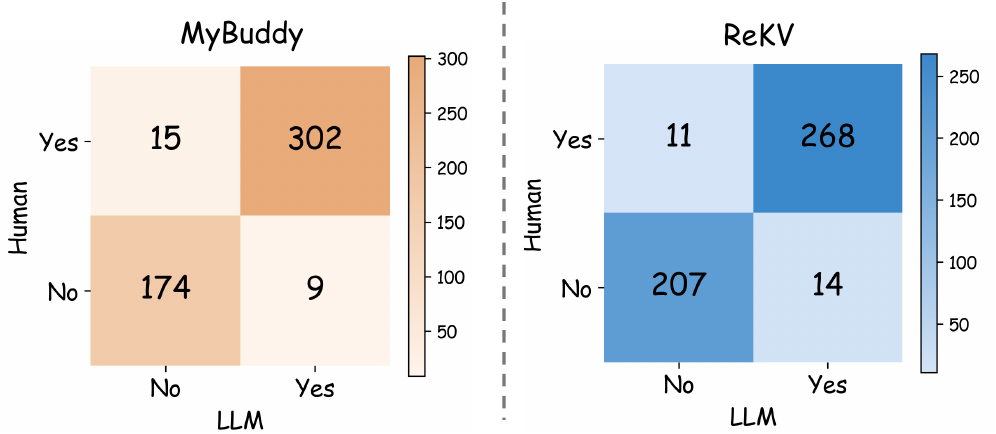}
    \caption{Confusion matrices of evaluation results.}
    \label{fig:supp_heatmap}
    \vspace{-4mm}
\end{figure}

Before actual model evaluation, we follow recent practices~\citep{cheng2024egothink,xiao2025egoblind} to tune the evaluation prompt to maximally align the LLM evaluator's judgment with that of humans. 
Specifically, we apply the offline general VideoQA and online streaming VQA methods (InternVL-3.5~\citep{wang2025internvl3} and ReKV~\citep{di2025streaming}, respectively) on 500 randomly selected samples from \benchname. We then invite four human volunteers to assess the model predictions by comparing them with the annotated ground-truth answers while watching the corresponding videos. Human evaluations are then aggregated following a majority voting mechanism. Afterwards, we tune the evaluation prompt for the LLM evaluator GPT-5~\citep{GPT-5}, and tune the prompt so that the LLM’s judgments align as closely as possible with those of humans.
According to the results in Tab.~\ref{tab:llm-as-judge}, GPT-5 with the final adjusted prompt and human evaluators achieve similar accuracy and score, with Pearson correlation coefficients of 0.87 and 0.91, respectively, indicating strong consistency. The Cohen’s kappa coefficient among the four volunteers is 0.81 for accuracy, indicating a high level of human agreement. Fig.~\ref{fig:supp_heatmap} further shows that the LLM evaluator exhibits a high level of consistency with human judgments, as reflected by the dominant mass along the agreement diagonal. Both positive and negative decisions display strong alignment, indicating that the model reliably mirrors human preferences when assessing semantic correctness. 
\begin{table}[t!]
\centering
\small
\caption{Comparison of human and LLM evaluators in terms of accuracy and score. The results show highly aligned scores between human and LLM judge.}
\setlength{\tabcolsep}{8pt}
\begin{tabular}{lcccc}
\toprule
\multirow{2}{*}{\textbf{Evaluator}} & \multicolumn{2}{c}{\textbf{InternVL-3.5}} & \multicolumn{2}{c}{\textbf{ReKV}} \\
\cmidrule(lr){2-3} \cmidrule(lr){4-5}
 & \textbf{Acc.(\%)} & \textbf{Sco.} & \textbf{Acc.(\%)} & \textbf{Sco.} \\
\midrule
Human & 62.3 & 3.1 & 55.3 & 2.9 \\
GPT-5 & 61.8 & 3.1 & 53.7 & 2.8 \\
\bottomrule
\end{tabular}
\vspace{-0.6cm}
\label{tab:llm-as-judge}
\end{table}

%% file: conclusion.tex
\section{Conclusion}
In this paper, we introduced a novel benchmark, \benchname, towards companion-style video question answering in first-person vision. \benchname\ stands out from traditional offline VideoQA and streaming VQA by capturing two key insights in companion-style assistance: Interactive Chaining and Ego-Deictics. Moreover, the QAs are timestamp-specific and are well grouped into four assistance types for better analysis.
We also proposed a novel companion-style QA framework \modelname, which features multi-level memory and cross-modal chain-of-thought reasoning for efficiently memorizing video representations in a streaming setting and interpreting the user’s actual intent from multi-modal historical contexts, respectively. \modelname\ achieves superior performance over strong competitors on both \benchname\ and existing long streaming VQA benchmarks, suggesting its high effectiveness. The ablations further demonstrate the contribution of our innovative designs. We hope that our dataset, model, and analysis could advance MLLMs towards companion-style QA in first-person vision.   

%% file: limitation.tex
\section{Limitations and Future Work}
\textit{Asynchronous Memory Consolidation.}
The Low-Active Memory summarization is designed to run during idle intervals, allowing the system to compress historical visual information without interfering with real-time QA responses. This design implicitly assumes that the user is not continuously issuing queries. Yet in highly interactive scenarios where questions arrive in rapid succession, opportunities for such background consolidation may become limited, potentially introducing latency if summarization must be executed alongside the QA pipeline. 
Our future work will explore more efficient or incremental memory consolidation strategies that operate continuously with minimal computational overhead.

\textit{Multi-user Assumption in Egocentric Interaction.}
Currently, \benchname\ assumes a single primary user interacting with the assistant. In realistic environments, however, egocentric scenes may contain multiple people or agents, making referential interpretation more complex. While our current ego-deictic grounding relies primarily on cues such as gaze direction and hand interactions, richer multi-agent reasoning may be required to disambiguate references in more complex social contexts.
Our future research would extend companion QA systems to multi-user scenarios by incorporating person tracking, speaker-aware dialogue modeling, and social interaction understanding, enabling assistants to reason jointly over visual observations and conversational participants.

%% file: appendix.tex
\clearpage

\begin{appendices}
\section{\benchname~Dataset}
\subsection{Video Activities}

\begin{figure}[ht!]
    \centering
    \includegraphics[width=\linewidth]{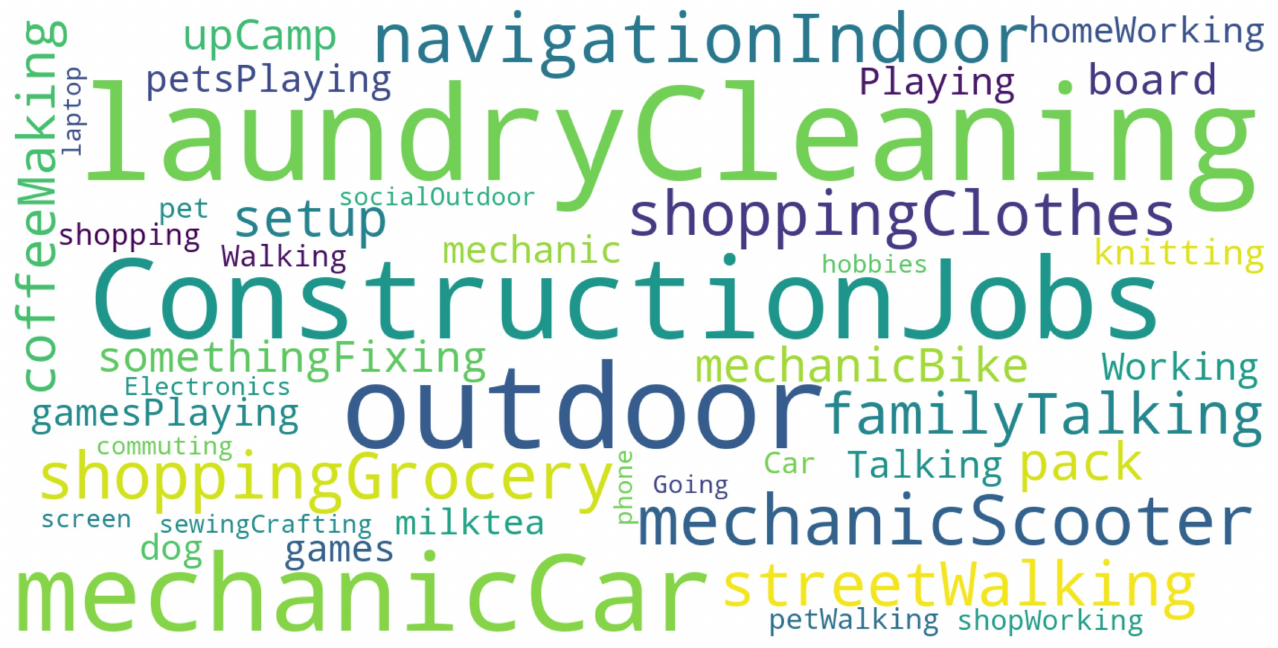}
    \caption{Activities in \benchname. }
    \label{fig:activities}
\end{figure}

To construct our dataset, we selected 1,012 videos from Egoschema~\citep{mangalam2024egoschema}, sourced from Ego4D~\citep{grauman2022ego4d}, which contains more than three thousand hours of worldwide daily-life recordings across diverse scenarios such as household, outdoor, workplace, leisure, and construction. These selected videos span a broad spectrum of indoor and outdoor daily activities (as shown in Fig.~\ref{fig:activities}) that are common in assistive scenarios.

\subsection{Automatic Question Candidate Generation}
Tab.~\ref{tab:supp_pro_generation} shows the prompts used for automatic question generation. After generation, we apply an additional filtering stage to remove irrelevant, redundant, or trivially answerable questions. Specifically, we prompt Gemini-2.5-pro~\citep{comanici2025gemini} to (1) evaluate the semantic relevance of each question to its corresponding video segment, (2) detect near-duplicate phrasings within the same highlight moment, and (3) reject questions whose answers can be easily inferred without visual information.

\subsection{Details of Human Annotation}
We recruit 25 volunteers from diverse countries and social backgrounds (students, engineers, doctors, freelancers, accountants, insurance agents, etc.) via the \emph{Prolific} platform to annotate QA pairs based on the automatically generated highlight moments and draft questions.

We clarify several key points of the human question–answer annotation process:
\begin{enumerate}
    \item We provide annotators with detailed definitions of all question categories to minimize ambiguity. For each category, we also supply several examples (as shown in Tab.~\ref{tab:category}) to guide question formulation.
    \item In the streaming setting, annotators are strictly restricted to viewing only the video content \textit{before} the ending timestamp of the highlight moment. Likewise, answers must rely solely on information available up to that same timestamp.
    \item Even for subjective questions, each answer must be evidence-based and grounded exclusively in the visual content observed before the highlight’s end time.
    \item To encourage natural and conversational phrasing, annotators are allowed to use ego-deictic expressions as well as abbreviations, slang, and other colloquial forms.
    \item We define interactively chained questions as those closely related to the prior QA pair within the same highlight moment. However, we do not require subsequent questions to share the same subject or phrasing. We also prevent chains from looping by disallowing any chained question from becoming the base question of another chain.
    \item Annotators are prohibited from using AI tools during the annotation process. Additionally, they must ensure that at least 50\% of their annotated QA pairs are fully written by themselves, independent of the draft.
\end{enumerate}

Overall, we establish clear guidelines to ensure that human-authored questions and answers are natural, temporally grounded, and well-defined, enabling the construction of a realistic companion QA dataset in an egocentric setting.

\begin{table*}[t!]
    \centering
    \caption{Question categories and examples of \benchname.}
    \resizebox{0.65\linewidth}{!}{%
    \begin{tabular}{l|c|l}
    \toprule
    \textbf{Category} & \textbf{Q Num. \& Ratio }& \makecell[c]{\textbf{Examples}}\\
    \hline
    \cellcolor{lightpurple!25}\textbf{Recall}  & 5631(26.1\%)  & \makecell[l]{Who just passed by me? \\ How much time passed after I started running? \\ What was he doing several seconds ago? \\ Did I lock the door before leaving? } \\
    \hline
    \cellcolor{lightblue!25}\textbf{Recognition} & 5069(23.5\%) & \makecell[l]{What's on that? \\What’s that red thing next to the greens? \\Who did I just hand those to?} \\
    \hline
    \cellcolor{lightorange!25}\textbf{Scrutinization} & 5523(25.6\%) & \makecell[l]{Am I missing any spot while sweeping? \\ Am I doing this fast enough? \\ Did I take longer to complete this than before? \\ Are these clothes of good quality?  }\\
    \hline
    \cellcolor{lightgreen!25}\textbf{Advisory} & 5351(24.8\%) & \makecell[l]{ What should I do next? \\ How long will it take to get through this line? \\ How do I know when the area is smooth enough? }\\
    \bottomrule
    \end{tabular}
    }
    \label{tab:category}
\end{table*}

\subsection{Interactively Chaining Analysis}
We identify three primary types of chained relationships:
\begin{itemize}
\item \textit{Referential chains} (41.2\%), where follow-up questions use pronouns or vague expressions that depend on previous answers for grounding (e.g., “it,” “there,” “that”);
\item \textit{Temporal chains} (23.5\%), where reasoning relies on earlier events or actions (e.g., “What did I do before this?”);
\item \textit{Inferential chains} (21.8\%), where subsequent questions seek higher-level interpretation or evaluation based on prior answers (e.g., asking “Did I do it correctly?” after “What did I just do?”).
\end{itemize}

Referential chains primarily evaluate a model’s ability to resolve ambiguity by grounding pronouns or underspecified expressions to the correct visual or contextual entities, requiring strong multimodal alignment and memory retrieval. Temporal chains demand coherent tracking of event sequences over time, including robust temporal indexing and the capacity to infer continuity or causality across past observations. Inferential chains require even deeper abstraction: in addition to recalling earlier answers, the model must reason about their implications, evaluate correctness, or draw higher-level conclusions.
Collectively, these chain types progress from surface-level referent grounding to increasingly deep temporal, causal, and evaluative reasoning, highlighting the cognitive and computational challenges in producing coherent responses for companion QA.

\section{\modelname\ Investigations and Experiments}

\subsection{Study of \modelname\ Design}

\textbf{Chained and Unchained Questions.}
Before designing our companion QA framework, we conducted a series of empirical studies to understand how question dependency influences both performance and computational efficiency in egocentric streaming settings. In particular, we investigated whether questions should always be treated as dependent on previous QA turns or as entirely independent.

Our findings show that treating all questions as dependent introduces substantial context contamination: accumulated historical QA injects irrelevant or distracting information that interferes with grounding the current visual scene. This practice not only reduces accuracy from 65.1\% to 56.5\%, but also increases inference time by 32.7\%. In contrast, treating all questions as independent loses essential contextual cues, leading to a sharp overall accuracy drop from 65.1\% to 38.2\% due to missing referential anchoring.

We then explored whether combining partial context (specifically, working memory and episodic memory) could alleviate these issues. Results indicate that leveraging both of them and further expanding to a multi-level memory (active, semi-active, low-active) module yields a more favorable balance between efficiency and contextual completeness. Finally, we evaluated different question-routing strategies and found that a lightweight zero-shot LLM prompting method outperformed a trained binary classifier (86.7\% \vs 73.5\%) with only a modest 12.5\% increase in inference time.
These observations underscore the need for an adaptive mechanism that dynamically determines whether a question is dependent or independent and selectively retrieves the appropriate context, motivating the design of our hierarchical reasoning model.

\begin{table}[t!]
\centering
\small
\caption{Comparison of disambiguation accuracy (D Acc.) and latency w/ and w/o hierarchical rephrasing.}
\label{tab:disambig_acc}
\resizebox{\linewidth}{!}{%
\begin{tabular}{lcc}
\toprule
\textbf{Method} & \textbf{D Acc.(\%)} & \textbf{Latency} \\
\midrule
Direct Question Answering & 62.4 & 423ms\\
Hierarchical Rephrasing (Ours) & \textbf{86.7} & 552ms \\
\bottomrule
\end{tabular}
}
\vspace{-2mm}
\end{table}

\textbf{Hierarchical Rephrasing.}
The hierarchical rephrasing module applies a chain-of-thought idea to explicitly resolve deictic pronouns and vague linguistic expressions before answer generation, thereby improving referent grounding. As shown in Tab.~\ref{tab:disambig_acc}, this approach improves disambiguation accuracy (measured as the proportion of rephrased questions whose referents are correctly grounded in the scene) while introducing limited latency. Empirical results indicate a substantial gain from 62.4\% to 86.7\% over direct prompting, demonstrating that hierarchical rephrasing effectively mitigates linguistic ambiguity with minimal additional inference latency. Specifically, by incorporating both gaze direction and recent hand–object interactions as embodied features, and utilizing the filtered visual information from multi-level memory, the model identifies and replaces deictic terms with their most likely visual referents by doing hierarchical rephrasing.

We further analyze different types of deictic situations, including unclear referents, vague temporal/spatial expressions, and ambiguous actions (see Tab.~\ref{tab:rephrasing_examples}). Questions containing ego-deictics exhibit the largest relative improvement, highlighting the module’s strength in resolving underspecified referents. Meanwhile, vague temporal/spatial and action-related questions also benefit from clearer contextual grounding that supports more reliable temporal and procedural reasoning. In general, the hierarchical reformulation process reduces misgrounded references by roughly 24\%, underscoring its ability to convert linguistically ambiguous questions into explicit, interpretable queries that improve both referent clarity and downstream answer correctness.

\begin{table*}[t!]
\centering
\small
\caption{Examples of hierarchical input rephrasing of \modelname.}
\label{tab:rephrasing_examples}
\resizebox{\linewidth}{!}{
\begin{tabular}{p{3cm}p{3.3cm}p{6.3cm}p{3.5cm}}
\toprule
\textbf{Deictic Type} & \textbf{Original Question} & \textbf{Hierarchically Rephrasing} & \textbf{Final Answer} \\
\midrule
Unclear referents & ``I need more, where is it?'' & ``What does \textit{it} refer to?'' $\rightarrow$ ``Does \textit{it} refer to the cup?'' $\rightarrow$ ``Where is the cup?'' & ``On the table.'' \\
\addlinespace[2pt]
Vague temporal / spatial cues & ``Has it started yet?'' & ``Does \textit{it} refer to the meeting?'' $\rightarrow$ ``Has the meeting started yet?'' & ``Yes, 2 minutes ago.'' \\
\addlinespace[2pt]
Ambiguous actions & ``Am I doing this right?'' & ``Does \textit{this} refer to cutting the onion?'' $\rightarrow$ ``Am I cutting the onion correctly?'' & ``Yes, but hold the knife away from your hand.'' \\
\bottomrule
\end{tabular}
}
\end{table*}

\begin{table}[t!]
\centering
\small
\vspace{-0.4cm}
\caption{Accuracy of eye gaze points and hand pose detection, and the correlation of correctly detected features with questions in terms of correlated percentage (Corr.) and score.}
\label{tab:gaze_hand_raw}
\resizebox{\linewidth}{!}{%
\begin{tabular}{lccc}
\toprule
\multirow{2}{*}{\textbf{Method}} & \multirow{2}{*}{\textbf{Accuracy(\%)}} & \multicolumn{2}{c}{\textbf{Corr. with Ques.}} \\
\cmidrule(lr){3-4}
&& \textbf{Corr.(\%)}& \textbf{Score} \\
\midrule
Human & 76.2 & 82.3 & 3.8 \\
GPT-5~\citep{GPT-5}   & 81.3 & 89.1 & 4.1\\
\bottomrule
\end{tabular}
}
\end{table}
\textbf{Embodied Ego-cues.}
To further evaluate the effectiveness of detecting embodied visual features for companion QA, we conduct both automatic and human assessments to determine whether the detected coordinates of eye-gaze points and hand-pose bounding boxes provide meaningful support for question answering. We randomly sample 500 samples from \benchname\ and recruit three human evaluators, using GPT-5~\citep{GPT-5} as an automatic evaluator. As presented in Tab.~\ref{tab:gaze_hand_raw}, human assessments show that 76.2\% of the detections are correctly anchored to the user’s eye gaze or hand poses, and of them 82.3\% are judged to be closely related to the corresponding questions and beneficial for answer generation.
These results demonstrate that (1) embodied feature extraction meaningfully enhances a model’s ability to interpret visual context during reasoning, and (2) filtering detected signals based on the incoming question further improves the relevance of the extracted features to the task.

\begin{table}[t!]
\centering
\small
\caption{Controlled ablation study on Buddy.}
\resizebox{\linewidth}{!}{
\begin{tabular}{lccc}
\toprule
\textbf{Configuration} & \textbf{Overall Acc.} \\
\midrule
Base GPT-5.5 Backbone (Raw F+His. QAs) & 50.5 \\
Buddy w/o Memory + Q Filter + Visual Info + Rephrasing & 52.2 \\
Buddy w/o Memory + Q Filter + Visual Info  & 56.1 \\
Buddy w/o Memory + Q Filter & 59.2 \\
Buddy w/o Memory & 62.4 \\
\midrule
Full \modelname (GPT-5.5) & \textbf{65.1} \\
\bottomrule
\end{tabular}
}
\label{tab:gpt5_control}
\end{table}
\textbf{Controlled Ablation Study.}
Tab.~\ref{tab:gpt5_control} further discerns the performance gains over the GPT-5.5~\citep{GPT-5.5} backbone by modules. The results demonstrate that each component of Buddy contributes progressively to the final performance. Starting from the GPT-5.5 backbone with raw frames and historical QA context (50.5\%), adding visual information retrieval and hierarchical rephrasing improves accuracy to 52.2\%. Introducing the Question Filter further increases performance to 56.1\%, indicating the importance of selecting relevant conversational context rather than using the entire QA history. Combining all components achieves the best performance of 65.1\%, demonstrating that the gains arise from the complementary effects of memory, context selection, and ego-deictic reasoning rather than from the GPT-5 backbone alone.

\subsection{Details of Baseline Models}
We provide a concise introduction to the MLLMs evaluated in experiments. 

\begin{itemize}
    \item \textbf{GPT-5.5~\citep{GPT-5.5}} not only outperforms previous GPT models on benchmarks and answers questions more quickly, but is more useful for real-world queries.
    \item \textbf{Gemini-3.1-pro~\citep{Gemini3.1}} surpasses previous versions as Google's most advanced ``thinking'' model, leveraging enhanced native reasoning to achieve state-of-the-art performance in complex coding, mathematical, and multimodal tasks.
    \item \textbf{Qwen3.5~\citep{qwen3.5}} is a recent native multimodal model family developed by the Qwen Team, targeting multimodal agent scenarios that require joint visual understanding, reasoning, coding, and interaction. Qwen3.5 enables efficient native multimodal training via a heterogeneous infrastructure that decouples parallelism strategies across vision and language components, avoiding uniform approaches’ inefficiencies. By exploiting sparse activations for cross-component computation overlap, it achieves near 100\% training throughput versus pure-text baselines on mixed text-image-video data.
    \item \textbf{LongVA~\citep{zhang2024long}} uses Qwen2-7B-Instruct~\citep{wang2024qwen2} as the backbone language model and performs pretraining with a context length of 224K over a total of 900M tokens. LongVA improved the capability of LLMs to handle long videos thanks to the long context transfer phenomenon.
    \item \textbf{InternVL-3.5~\citep{wang2025internvl3}} follows the “ViT–MLP–LLM” paradigm adopted in previous versions of InternVL. InternVL-3.5 initializes the language model using the Qwen3 series~\citep{bai2025qwen3} and GPT-OSS~\citep{GPT-OSS}, and the vision encoder using InternViT-300M and InternViT-6B~\citep{chen2024internvl}. The Dynamic High Resolution strategy in InternVL1.5~\citep{chen2024far} is also retained. InternVL3.5 adopts a new reinforcement learning framework Cascade RL, together with two new modules Visual Resolution Router and Decoupled Vision-Language Deployment, to reduce the inference cost.
    \item \textbf{VideoChat2~\citep{li2024mvbench}} applies UMT-L~\citep{li2023unmasked} as visual encoder and Vicuna-7B v0~\citep{Vicuna} as LLM, and deploy QFormer using the pretrained \(BERT_{base}\) ~\citep{devlin2019bert}. It performs a progressive multi-modal training strategy with 3 stages.
    \item \textbf{LLaMA-VID~\citep{li2024llamavid}} instantiates the model with the pre-trained EVA-G~\citep{fang2023eva} for visual encoder and QFormer~\citep{dai2023instructblip} for text decoder. It builds the instruction pairs from~\citep{liu2024improved} and ~\citep{maaz2023video} for instruction tuning, and builds long QA pairs from movies and corresponding scripts in MovieNet~\citep{huang2020movienet} for training. LLaMA-VID designed a dual-token paradigm to empower existing LLMs to support long videos. 
    \item \textbf{Video-LLaMA3~\citep{zhang2025videollama}} applies Qwen2.5-7B~\citep{bai2025qwen2} as LLM initialization, and the vision encoder is initialized with the fine-tuned SigLIP~\citep{zhai2023sigmoid} weights. It uses a variety of scene images and document recognition images, along with a small portion of scene text images for training, and collects videos with generally annotated captions, questions, and answers from multiple open-source datasets for video-centric fine-tuning.
    \item \textbf{Video-XL~\citep{shu2025video}} is trained on Qwen-2-7B ~\citep{wang2024qwen2}. During pre-training, Laion-2M dataset~\citep{he2024efficient} is used to optimize the projector, where visual embeddings from CLIP-ViT-L~\citep{radford2021learning} based vision encoder are aligned with the text embeddings of LLM. During fine-tuning, visual instruction tuning is applied to optimize the parameters of the vision encoder, projector, and LLM. 
    \item \textbf{LVAgent~\citep{chen2025lvagent}} uses a constructed agent library. The ASP-CLIP~\citep{chen2023attentive} model, optimized using Adam~\citep{kingma2014adam}, is used as the retrieval model for temporal context modeling.
\end{itemize}
\begin{itemize}
    \item \textbf{Cogstream~\citep{zhao2025cogstream}} chooses VideoLLaMA3~\citep{zhang2025videollama} as baseline, comprising VL3-SigLIP-NaViT~\citep{zhai2023sigmoid} as video encoder, an MLP projection layer, and Qwen2.5~\citep{bai2025qwen2} as language model. Cogstream designs a reasoner to compress the accumulated video stream and reason over the integrated visual-textual information.
    \item \textbf{Dispider~\citep{qian2025dispider}} utilizes CLIP-L/14~\citep{radford2021learning} as frame encoder, Qwen2-7B~\citep{wang2024qwen2} as final LLM, and Qwen2-1.5B~\citep{wang2024qwen2} to produce time-aware compressed clip-wise features. Dispider features a lightweight Proactive Streaming Video Processing module that tracks the video stream and identifies optimal moments for interaction.
    \item \textbf{Flash-VStream~\citep{zhang2024flash}} performs initialization of visual encoder, projector, and LLM from pretrained Qwen2-VL-7b~\citep{wang2024qwen2}, and adapts a 9k subset from LLaVA-Video dataset~\citep{zhang2024video} to do a LoRA~\citep{hu2022lora} instruction tuning. Flash-VStream designs a Flash Memory module that retrieves detailed spatial information.
    \item \textbf{ReKV~\citep{di2025streaming}} integrates LLaVA-OV-7B~\citep{li2024llava} for evaluation. ReKV is a training-free approach that stores processed video in key-value caches (KV-Caches) and also introduces a retrieval method that retrieves only query-relevant KV-Caches.
\end{itemize}

\section{Model Prompts}
Tab.~\ref{tab:supp_pro_generation} provides the prompts used by Gemini-2.5-pro for highlight moment detection, question generation and filtering. 
Tab.~\ref{tab:supp_liveqa_eval} lists the prompts employed by GPT-5 for model evaluation. Tab.~\ref{tab:supp_pro_summarization} shows the prompts used in Low-Active Memory of \modelname\ for summarization.
Tab.~\ref{tab:supp_pro_question_filter} details the specific prompts applied for the Question Filter in~\modelname.  
Tab.~\ref{tab:supp_pro_vir} includes the prompts designed for the visual information retrieval step with multi-level memory inputs.
Tab.~\ref{tab:supp_pro_qa} further illustrates the prompt for the final zero-shot question answering with hierarchical rephrasing.

\begin{table*}[t!]
\caption{Prompts for automatic question candidate generation of \benchname.}
\centering
\label{tab:supp_pro_generation}
\fontsize{9}{10}\selectfont 
 \resizebox{1.0\textwidth}{!}{
\begin{tabular}{p{17cm}}
\hline
\makecell[c]{\textbf{Automatic Question Candidate Generation Prompts}}  \\
\hline
\# Highlight detection:\\
You are an advanced assistant specialized in video analysis. \\
Your task is to analyze the provided video and identify its key highlights based on significant visual content, such as rapid action, noteworthy events, or shifts in visual dynamics. The input is a video file and the output should be a list of timestamps that correspond to the start and end of each highlight segment in the video. Please focus on identifying segments with high visual or contextual importance. Answer with ONLY timestamp pairs. \\
Ensure the timestamps are accurate and clearly marked in seconds, e.g., {[00:30, 00:45], [02:21, 02:51]}. \\
Video: \emph{\{video\}} \\
\hline
\# Question generation: \\
You are a good question generator, please help me generate appropriate questions that may be asked by users of an embodied visual companion, given an egocentric video and some highlight moments. \\
Task instructions for each highlight moment: \\
1. Describe Actions: First summarize the key action/event in a phrase, then clearly detail what the main characters are doing in each highlight moment, focusing on their actions and interactions. Refer to the camera wearer as 'c'. \\
2. Generate Questions and Answers: Generate questions  for each highlight moment according to the following requirements:\\
Requirement 1: Egocentric View: Ask the question with the assumption that you are the camera wearer and are new to this activity.\\
Requirement 2: Challenge-Specific: Reflect various challenges faced by different identities in distinct scenes. \\
Requirement 3: Diversity: Ensure questions are varied and explore different facets of the video content, and try to make some questions correlated with each other.\\
Requirement 4: IMPORTANT: Use Pronouns: Employ at least one pronoun like I, you, we, he, she, them, it, etc. in the question, rather than specific names and objects. \\
Here is an example: \\
Highlight Moment 1: [00:00, 00:20]\\   
Action Description:\\
Extending cup: C is extending a cup to another individual who is standing across the table.\\
Questions and Answers:\\        
1.  Question:  Is it full?\\        
Answer: Yes, the cup is full with coffee.\\    
2.  Question:  Where can I put it after I'm done?\\         
Answer: You can put it on the table.\\    
3.  Question:  I am tired, how can we enjoy it?\\     
Answer: You can sit near the table and enjoy the drink with another person.\\    
4.  Question: What's he wearing?\\        
Answer: The man next to you is wearing a blue T-shirt.\\    5.  Question:  What is the next step for me once I finish this?\\         
Answer: You will need to go to the kitchen and wash the cup.\\ 
Other example questions could be: What was I doing 10 seconds ago? Is this edible? Where is it? How long will it take to get through this? What does that road sign say? What should I do next? Who did I hand the chopsticks to? Why should I do this? Where is my key? How do I use the medication in this bottle?\\
Video: \textit{\{video\}} Highlight moments: \textit{\{highlight\}} \\
\hline
\# Automatic question filtering:\\
You are a helpful assistant. Given a set of questions and their corresponding video, evaluate whether these questions are semantically relevant to the visual content, detect if they are redundant or highly similar to any other, and determine whether they can be trivially answered without meaningful temporal grounding.\\
Video: \textit{\{video\}} Questions: \textit{\{question\}} \\
\hline
\end{tabular}
}
\end{table*}

\begin{table*}[t!]
\caption{Prompts for GPT-5~\citep{GPT-5} to evaluate MLLMs on~\benchname.}
\centering
\label{tab:supp_liveqa_eval}
\fontsize{9}{12}\selectfont
 \resizebox{1.0\textwidth}{!}{
\begin{tabular}{p{17cm}}
\hline
\makecell[c]{\textbf{Evaluation Prompts for~\benchname}}  \\
\hline
You are an intelligent chatbot designed for evaluating the correctness of generative outputs for egocentric, streaming question-answer pairs in the BuddyVQA dataset.  
Your task is to compare the predicted answer with the correct answer set and determine whether they meaningfully match, taking into account the characteristics of egocentric video, ego-deictic expressions, and interactive question chains. \\[4pt]
\\
The following are the requirements of evaluation: \\
a. Focus on the meaningful match between the predicted answer and the correct answer sets.  
In BuddyVQA, many questions contain ego-deictic expressions such as ``this", ``that", ``it", ``here", or ``there". Ensure that the predicted answer resolves the same referent as the ground truth, even if expressed differently. \\
b. The correct answers are provided as a \emph{set}.  
Your evaluation should treat the prediction as correct if it meaningfully matches \emph{any} of the answers in the set.  
The match does not need to be exact, but must refer to the same grounded entity, event, or semantic content. \\
c. Consider the temporal continuity of chained questions.  
Some BuddyVQA questions depend on prior context (\eg follow-up questions like ``Is it still there?" or ``Did I do it correctly?").  
When evaluating, ensure that the predicted answer maintains the same contextual interpretation and refers to the correct entity or past event, consistent with the ground-truth chain. \\
d. Synonyms, paraphrases, or alternative phrasings are valid as long as they preserve the same meaning and refer to the same grounded visual entity or event. \\

\\
Provide your evaluation result only as a yes/no and score where the score is an integer value between 0 and 5, with 5 indicating the highest meaningful match. \\
Please generate the response in the form of a Python dictionary string with keys 'pred' and 'score', where the value of 'pred' is a string of 'yes' or 'no' and the value of 'score' is an integer. For example, your response should look like this: \{'pred': 'yes', 'score': 5\}, \{'pred': 'no', 'score': 1\}.\\

Please evaluate the following video-based question-answer pair: \\
Question: \{\emph{question}\}  Correct Answer set: \{\emph{GT answer}\}  Predicted Answer: \{\emph{predicted answer}\} \\[2pt]
\hline
\end{tabular}}
\end{table*}

\begin{table*}[t!]
\caption{Prompt for zero-shot summarization model $\mathcal{S}$ used to construct Low-Active Memory.}
\centering
\label{tab:supp_pro_summarization}
\fontsize{9}{12}\selectfont
\resizebox{1.0\textwidth}{!}{
\begin{tabular}{p{17cm}}
\hline
\makecell[c]{\textbf{Low-Active Memory Summarization Prompt }} \\
\hline
You are an intelligent multimodal assistant tasked with summarizing egocentric video segments to extract long-term, stable, and reusable contextual information.  
Your goal is to produce a concise textual summary that captures information in long videos, including key activities, recurring patterns, background environment, salient objects, user routines, and consistent spatial layout, etc.\\
\\
\# Instructions: \\
- Focus on information that remains stable or frequently appears across this time window. Identify recurring actions, repeated interactions, or common tasks performed by the user. \\
- Capture stable environmental structure such as room layout, tool locations, furniture arrangement, or habitual user workflows. \\
- Extract objects or entities that reliably appear or are frequently manipulated across the window. \\
\\
Your summary should represent general knowledge useful for future reasoning, not for a specific question. Write the summary in clear, compact sentences that capture high-level patterns and omit unnecessary detail. Provide only the long-term summary text without explanation or extra commentary. \\
Please summarize the following egocentric video: 
\{\emph{video features}\} \\
\hline
\end{tabular}
}
\end{table*}

\begin{table*}
\caption{Prompt for the Question Filter $\mathcal{R}$ used to detect chained questions and retrieve relevant historical QA pairs.}
\centering
\label{tab:supp_pro_question_filter}
\fontsize{9}{12}\selectfont
\resizebox{1.0\textwidth}{!}{
\begin{tabular}{p{17cm}}
\hline
\makecell[c]{\textbf{Question Filtering Prompt}} \\
\hline
You are an intelligent multimodal assistant designed to identify whether an incoming question in a streaming egocentric QA scenario is \emph{chained} (dependent on prior QA turns) or \emph{unchained} (independent).  
Your task is to analyze the current question, compare it with a buffer of historical QA pairs and the current frame, and determine:  
(1) whether the question relies on earlier context, and  
(2) if so, which prior QA pair is the most relevant for resolving this dependency. \\[4pt]
\\
\# Instructions: \\
- Classify the incoming question as either \emph{``chained''} or \emph{``unchained''}.  
A chained question typically contains ego-deictic expressions (e.g., ``this'', ``that'', ``it'', ``there'') or implicit references to past events, actions, or answers (e.g., ``Is it still there?'', ``Did I do it correctly?'', ``What about now?'') that require information from historical QAs to answer.  
An unchained question is self-contained and understandable without referencing earlier context. \\

- If the question is classified as unchained, return the label ``unchained'' and do not select any historical QA pair. \\

- If the question is classified as chained, inspect the historical QA buffer and identify the single most relevant QA pair that resolves the ambiguous referent or contextual dependency.  
Relevance may arise from: semantic similarity, shared objects or actions, temporal continuity, or matching referential expressions. \\

- Focus only on meaningful contextual grounding, not on superficial word overlap.  
Your output must identify the QA pair that best supports resolving the question’s dependency. \\
\\
Provide the final result strictly in the following Python dictionary format:  
\{'type': 'chained' or 'unchained',  
\ \ 'base\_qa': qid of the most relevant QA pair or \texttt{null} if unchained\}. 
Do not include any explanation, reasoning steps, or additional text. \\[6pt]
Please analyze the incoming question: \\
Current Question: \{\emph{question}\} Historical QAs: \{\emph{historical QAs}\} Current frame: \{\emph{frame features}\} \\[4pt]
\hline
\end{tabular}
}
\end{table*}

\begin{table*}
\caption{Prompt for the Visual Information Retrieval (VIR) module.}
\centering
\label{tab:supp_pro_vir}
\fontsize{9}{12}\selectfont
 \resizebox{1.0\textwidth}{!}{
\begin{tabular}{p{17cm}}
\hline
\makecell[c]{\textbf{Visual Information Retrieval Prompt}} \\
\hline
You are an intelligent multimodal assistant designed to retrieve visual information from egocentric video memories in order to support real-time question answering.  
Your task is to (1) generate concise visual captions, (2) filter long-term summaries for information relevant to the incoming question, and (3) detect embodied features such as eye-gaze points and hand bounding boxes that help ground deictic references.  
All three subtasks must be performed using the same lightweight MLLM capabilities. \\[6pt]
\\
\# Instructions for Captioning (from Active and Semi-Active Memory): \\
- Generate a concise and accurate caption describing the visually salient content in the provided video frames.  
Focus on the objects, spatial relations, user interactions, and scene attributes that are relevant for answering a question.  \\
- The caption should reflect only the visible content and be grounded in the egocentric viewpoint. \\[6pt]
\\
\# Instructions for Memory Filtering (from Semi-Active and Low-Active Memory): \\
- Given the incoming question, identify memory descriptions that are semantically or visually relevant.  
Relevance may arise from object continuity, spatial layout, repeated actions, or referents implied by ego-deictic expressions such as ``this'', ``that'', ``it'', or ``there''.  \\
- Return only the long-term or intermediate descriptions that meaningfully support answering the question.  
Do not repeat irrelevant or outdated information. \\[6pt]
\\
\# Instructions for Eye-Gaze and Hand-Pose Detection (from Active Memory): \\
- Detect eye-gaze fixation points and hand bounding boxes visible in the frames. Return their coordinates in a simple and interpretable format.  
- The detected features should help disambiguate the referent of the given question when the user employs deictic expressions or when the focus of attention is ambiguous.  
- Only output detections that are visually grounded and clearly identifiable; do not hallucinate features. \\[6pt]
\\
Provide the final result as a Python dictionary with the following keys:  
\{'caption': \emph{text}, 'filtered\_memory': \emph{text or list}, 'coordinates': \emph{list of gaze/hand coordinates}\}.  
Do not include any explanation or reasoning in your output.  
Only provide the dictionary. \\[6pt]

Please process the following inputs: \\
Incoming Question: \{\emph{question}\} Active Memory: \{\emph{$\mathcal{M}_A$}\} Semi-Active Memory: \{\emph{$\mathcal{M}_S$}\} Low-Active Memory: \{\emph{$\mathcal{M}_L$}\} \\[2pt]
\hline
\end{tabular}}
\end{table*}
\clearpage

\begin{table*}[t!]
\caption{Prompt for zero-shot question answering with hierarchical rephrasing.}
\centering
\label{tab:supp_pro_qa}
\fontsize{9}{12}\selectfont
 \resizebox{1.0\textwidth}{!}{
\begin{tabular}{p{17cm}}
\hline
\makecell[c]{\textbf{Zero-shot Question Answering Prompt}} \\
\hline
You are an intelligent multimodal assistant designed to answer egocentric, streaming questions using retrieved visual information, historical QA context, and embodied features.  
Your task is to first \emph{rephrase} the incoming question into an explicit, fully disambiguated form and then generate a correct answer grounded in the provided multimodal context. \\[4pt]
\\
\# Instructions for Hierarchical Input Rephrasing: \\
Rewrite the question in a clearer and explicit form that removes ambiguity.  
Resolve ego-deictic expressions such as ``this'', ``that'', ``it'', ``here'', or ``there'' by grounding them using the provided context.  
Incorporate historical QA, captions, and embodied features to determine the correct referent or event for rephrasing. 
The rephrased question must express the user’s intent explicitly and be fully interpretable without additional context.  
Provide only one rephrased question—concise, unambiguous, and grounded. \\[6pt]
Here is an example: \\

\textit{Incoming Question:} ``Where is it now?'' \\
\textit{Context Provided:}  
The historical QA buffer contains the pair (``Where did I put my phone?'', ``You placed your phone on the kitchen counter.'');  
the retrieved visual captions describe ``a phone on the counter near a bread box'';  
and the gaze coordinate indicates fixation on the counter area. \\[2pt]
\textit{Rephrased Question:} ``Where is the phone now?'' \\
\\

\# Instructions for Answering the Rephrased Question: \\
Using the rephrased question and the full multimodal context, generate a factually correct and contextually grounded answer.  
Ensure that the answer is consistent with visual evidence, spatial layout, object interactions, temporal history, and embodied cues (gaze and hand pose).  
Avoid hallucinating entities or events not supported by the provided context.  
The final answer should be concise, accurate, and specifically aligned with the rephrased question. \\[6pt]

\\
Your output must be a Python dictionary with the following fields: 
\{'rephrased\_question': \emph{text}, 'answer': \emph{text}\}.
Do not include explanation, reasoning traces, chain-of-thought, or additional commentary.  
Only provide the dictionary. \\[6pt]

Please process the following inputs: \\
Incoming Question: \{\emph{question}\} 
Context for Reasoning $\mathcal{C}_t$: \{
Retrieved Visual Descriptions: \{\emph{$\mathbf{V}^{des}_t$}\} ,
Relevant Historical QA: \emph{$(Q^{rel}_t, A^{rel}_i)$}, 
Embodied Feature Coordinates: \{\emph{$\mathbf{V}^{coor}_t$}\} \} \\[2pt]
\hline
\end{tabular}}
\end{table*}

\end{appendices}